\documentclass[10pt,twocolumn,letterpaper]{article}

\usepackage{cvpr}
\usepackage{cuted} %< for strip environment

\usepackage{currfile} %< for \currfilebase macro

\usepackage{caption} %< for "\captionof" commands (teaser)
\usepackage{booktabs}
\usepackage{multirow}
\usepackage{graphicx}
\usepackage{amsmath,amssymb,mathtools}

\definecolor{cvprblue}{rgb}{0.21,0.49,0.74}
\usepackage[pagebackref,breaklinks,colorlinks,allcolors=cvprblue]{hyperref}

\def\confName{}
\def\confYear{}

\newcommand{\DaThreeLBetaSmallShared}{0.32}

\newcommand{\DaThreeMetricLAlphaSmallShared}{\ensuremath{-0.05}}

\newcommand{\DaThreeMetricLCueContext}{0.79}

\newcommand{\DaThreeMetricLCueMagnify}{1.34}

\newcommand{\DaThreeMetricLCueZoom}{1.03}

\newcommand{\DaThreeNestedAlpha}{0.56}

\newcommand{\DaThreeNestedAlphaPlusBeta}{0.96}

\newcommand{\DaThreeNestedAlphaSmallShared}{0.50}

\newcommand{\DaThreeNestedBetaSmallShared}{0.40}

\newcommand{\DaThreeNestedCamGivenAlpha}{\ensuremath{-0.04}}
\newcommand{\DaThreeNestedCamGivenScale}{0.309}
\newcommand{\DaThreeNestedCamGivenScaleDiodeIn}{0.418}
\newcommand{\DaThreeNestedCamGivenScaleDiodeOut}{0.370}
\newcommand{\DaThreeNestedCamGivenScaleKitti}{0.301}
\newcommand{\DaThreeNestedCamGivenScaleNyu}{0.047}
\newcommand{\DaThreeNestedCamImages}{200}
\newcommand{\DaThreeNestedCamOwnAlpha}{0.56}
\newcommand{\DaThreeNestedCamOwnScale}{0.099}
\newcommand{\DaThreeNestedCamOwnScaleDiodeIn}{0.095}
\newcommand{\DaThreeNestedCamOwnScaleDiodeOut}{0.165}
\newcommand{\DaThreeNestedCamOwnScaleKitti}{0.195}
\newcommand{\DaThreeNestedCamOwnScaleNyu}{0.051}

\newcommand{\DaThreeNestedParams}{1,690}

\newcommand{\DaThreeNestedUnshrinkAlphacorr}{\ensuremath{-0.05}}
\newcommand{\DaThreeNestedUnshrinkBeta}{0.41}

\newcommand{\DavTwoBAlpha}{0.85}

\newcommand{\DavTwoBAlphaShared}{0.81}

\newcommand{\DavTwoLAlphaShared}{0.72}

\newcommand{\DavTwoSAbsrelThree}{0.652}
\newcommand{\DavTwoSAlpha}{0.95}
\newcommand{\DavTwoSAlphaHi}{1.12}
\newcommand{\DavTwoSAlphaLarge}{0.97}
\newcommand{\DavTwoSAlphaLo}{0.83}

\newcommand{\DavTwoSAlphaShared}{0.87}
\newcommand{\DavTwoSAlphaSmall}{0.85}

\newcommand{\DavTwoSBaseThree}{0.336}

\newcommand{\DavTwoSBilinear}{0.54}
\newcommand{\DavTwoSBlur}{1.3}

\newcommand{\DavTwoSBlurKittiAlpha}{0.82}

\newcommand{\DavTwoSBlurNyuAlpha}{0.56}

\newcommand{\DavTwoSCueContext}{0.96}

\newcommand{\DavTwoSCueMagnify}{0.58}

\newcommand{\DavTwoSCueShrink}{1.76}

\newcommand{\DavTwoSDigitalLr}{0.55}

\newcommand{\DavTwoSEftKittiAlpha}{0.18}
\newcommand{\DavTwoSEftKittiBlur}{29.6}

\newcommand{\DavTwoSEftNyuAlpha}{0.12}

\newcommand{\DavTwoSEftbDiodeInAlpha}{0.37}

\newcommand{\DavTwoSEftbDiodeOutAbsrelOne}{1.000}

\newcommand{\DavTwoSEftbDiodeOutAlpha}{0.40}

\newcommand{\DavTwoSEftbKittiAbsrelOne}{0.175}

\newcommand{\DavTwoSEftbKittiAlpha}{0.05}
\newcommand{\DavTwoSEftbKittiBlur}{4.9}

\newcommand{\DavTwoSEftbNyuAlpha}{0.07}
\newcommand{\DavTwoSEftbNyuBlur}{0.4}

\newcommand{\DavTwoSEftbSrrawOpticalAlpha}{0.59}

\newcommand{\DavTwoSEftbThreeKittiAbsrelOne}{0.173}

\newcommand{\DavTwoSEftbThreeKittiAlpha}{0.07}

\newcommand{\DavTwoSEftbThreeNyuAbsrelOne}{0.360}

\newcommand{\DavTwoSEftbThreeNyuAlpha}{0.05}

\newcommand{\DavTwoSFixDiodeInAbsrel}{0.365}

\newcommand{\DavTwoSFixDiodeInZoomed}{0.598}
\newcommand{\DavTwoSFixDiodeOutAbsrel}{0.661}
\newcommand{\DavTwoSFixDiodeOutTarget}{0.605}
\newcommand{\DavTwoSFixDiodeOutTargetMedian}{0.430}
\newcommand{\DavTwoSFixDiodeOutZoomed}{0.558}

\newcommand{\DavTwoSFixKittiAbsrel}{0.384}
\newcommand{\DavTwoSFixKittiTarget}{0.149}

\newcommand{\DavTwoSFixKittiZoomed}{0.758}
\newcommand{\DavTwoSFixNyuAbsrel}{0.495}

\newcommand{\DavTwoSFixNyuZoomed}{0.675}

\newcommand{\DavTwoSFixedDiodeInAbsrelOne}{0.280}

\newcommand{\DavTwoSFixedDiodeInAlpha}{0.70}

\newcommand{\DavTwoSFixedDiodeOutAbsrelOne}{0.731}

\newcommand{\DavTwoSFixedDiodeOutAlpha}{0.65}

\newcommand{\DavTwoSFixedKittiAbsrelOne}{0.138}

\newcommand{\DavTwoSFixedKittiAlpha}{1.03}
\newcommand{\DavTwoSFixedKittiBlur}{1.1}
\newcommand{\DavTwoSFixedNyuAbsrelOne}{0.344}

\newcommand{\DavTwoSFixedNyuAlpha}{0.98}

\newcommand{\DavTwoSJpeg}{1.00}
\newcommand{\DavTwoSJpegAfterZoom}{1.00}

\newcommand{\DavTwoSOptical}{0.58}

\newcommand{\DavTwoSRatioThree}{0.35}
\newcommand{\DavTwoSRegionDiff}{0.03}
\newcommand{\DavTwoSRegionObject}{1.07}
\newcommand{\DavTwoSRegionObjectLarger}{55}
\newcommand{\DavTwoSRegionP}{0.0088}
\newcommand{\DavTwoSRegionPairs}{429}
\newcommand{\DavTwoSRegionSurface}{1.01}

\newcommand{\DavTwoSScaleDiodeInAlpha}{0.23}

\newcommand{\DavTwoSScaleDiodeOutAlpha}{0.27}

\newcommand{\DavTwoSScaleKittiAlpha}{0.19}
\newcommand{\DavTwoSScaleKittiBlur}{7.3}
\newcommand{\DavTwoSScaleNyuAbsrelOne}{0.346}

\newcommand{\DavTwoSScaleNyuAlpha}{0.04}

\newcommand{\DavTwoSScalenbKittiBlur}{23.6}

\newcommand{\DavTwoSSrrawDigitalAlpha}{0.76}

\newcommand{\DavTwoSSrrawOpticalAlpha}{0.76}

\newcommand{\DavTwoSSrrawP}{0.45}

\newcommand{\DavTwoSUncAlphaLarge}{0.97}

\newcommand{\DavTwoSUncAlphaSmall}{0.85}

\newcommand{\DavTwoSZhongKittiAbsrelOne}{0.222}

\newcommand{\DavTwoSZhongKittiAlpha}{0.18}
\newcommand{\DavTwoSZhongKittiBlur}{25.8}

\newcommand{\DavTwoSZhongNyuAlpha}{0.11}

\newcommand{\DavTwoSZoomOfJpeg}{0.97}
\newcommand{\DavTwoSZoomOfJpegKitti}{0.86}
\newcommand{\DavTwoSZoomTwoFactor}{0.52}

\newcommand{\DepthproAlpha}{0.31}

\newcommand{\DepthproAlphaLarge}{0.32}

\newcommand{\DepthproAlphaPlusBeta}{0.97}

\newcommand{\DepthproAlphaSmall}{0.10}
\newcommand{\DepthproAlphaSmallShared}{0.23}

\newcommand{\DepthproBeta}{0.65}

\newcommand{\DepthproCamGivenAlpha}{\ensuremath{-0.03}}
\newcommand{\DepthproCamGivenScale}{0.189}
\newcommand{\DepthproCamGivenScaleDiodeIn}{0.358}
\newcommand{\DepthproCamGivenScaleDiodeOut}{0.816}
\newcommand{\DepthproCamGivenScaleKitti}{0.071}
\newcommand{\DepthproCamGivenScaleNyu}{0.063}

\newcommand{\DepthproCamOwnAlpha}{0.31}
\newcommand{\DepthproCamOwnScale}{0.136}
\newcommand{\DepthproCamOwnScaleDiodeIn}{0.138}
\newcommand{\DepthproCamOwnScaleDiodeOut}{0.458}
\newcommand{\DepthproCamOwnScaleKitti}{0.117}
\newcommand{\DepthproCamOwnScaleNyu}{0.064}

\newcommand{\DepthproCueMagnify}{0.84}

\newcommand{\DepthproCueShrinkFocal}{0.70}

\newcommand{\DepthproDigitalLr}{0.75}
\newcommand{\DepthproDsDiodeInAlpha}{0.19}
\newcommand{\DepthproDsDiodeOutAlpha}{\ensuremath{-0.01}}
\newcommand{\DepthproDsKittiAlpha}{0.35}
\newcommand{\DepthproDsNyuAlpha}{0.46}

\newcommand{\DepthproEftfFheadDiodeInAlpha}{0.44}

\newcommand{\DepthproEftfFheadDiodeOutAlpha}{0.44}

\newcommand{\DepthproEftfFheadKittiAlpha}{0.23}

\newcommand{\DepthproEftfFheadNyuAbsrelThree}{0.247}
\newcommand{\DepthproEftfFheadNyuAlpha}{0.09}

\newcommand{\DepthproFheadDiodeInAlpha}{0.19}

\newcommand{\DepthproFheadDiodeOutAlpha}{\ensuremath{-0.01}}

\newcommand{\DepthproFheadKittiAlpha}{0.35}

\newcommand{\DepthproFheadNyuAbsrelThree}{0.464}
\newcommand{\DepthproFheadNyuAlpha}{0.46}

\newcommand{\DepthproFocalThree}{2.05}

\newcommand{\DepthproJpegAfterZoom}{0.90}
\newcommand{\DepthproJpegAfterZoomFocal}{0.90}

\newcommand{\DepthproJzJpegOne}{1.00}

\newcommand{\DepthproOptical}{0.83}

\newcommand{\DepthproRegionDiff}{0.02}
\newcommand{\DepthproRegionObject}{0.38}
\newcommand{\DepthproRegionObjectLarger}{65}
\newcommand{\DepthproRegionP}{2.5e-13}

\newcommand{\DepthproRegionSurface}{0.36}
\newcommand{\DepthproSensAlphaAtHi}{0.13}
\newcommand{\DepthproSensAlphaAtLo}{\ensuremath{-0.09}}
\newcommand{\DepthproSensBetaCi}{0.70}
\newcommand{\DepthproSensBetaHi}{0.78}
\newcommand{\DepthproSensBetaLo}{0.62}

\newcommand{\DepthproSensMuShiftScaleOne}{0.245}
\newcommand{\DepthproSpreadRho}{0.00}
\newcommand{\DepthproSrrawDigitalAlpha}{\ensuremath{-0.08}}

\newcommand{\DepthproSrrawOpticalAlpha}{\ensuremath{-0.08}}

\newcommand{\DepthproUncBetaLarge}{0.65}

\newcommand{\DepthproUncBetaSmall}{0.83}

\newcommand{\DepthproUnshrinkAlpha}{0.31}
\newcommand{\DepthproUnshrinkAlphacorr}{0.04}
\newcommand{\DepthproUnshrinkBeta}{0.70}

\newcommand{\DepthproUnshrinkScaleOne}{0.136}
\newcommand{\DepthproUnshrinkScaleOnecorr}{0.138}
\newcommand{\DepthproUnshrinkScaleThree}{0.467}
\newcommand{\DepthproUnshrinkScaleThreecorr}{0.359}

\newcommand{\DepthproZoomOfJpeg}{0.82}
\newcommand{\DepthproZoomOfJpegFocal}{0.82}

\newcommand{\EftbSeedsDiodeInAbsrelOneMean}{0.430}
\newcommand{\EftbSeedsDiodeInAbsrelOneSd}{0.028}
\newcommand{\EftbSeedsDiodeInAlphaMean}{0.33}
\newcommand{\EftbSeedsDiodeInAlphaSd}{0.03}
\newcommand{\EftbSeedsDiodeOutAbsrelOneMean}{0.965}
\newcommand{\EftbSeedsDiodeOutAbsrelOneSd}{0.056}
\newcommand{\EftbSeedsDiodeOutAlphaMean}{0.41}
\newcommand{\EftbSeedsDiodeOutAlphaSd}{0.01}
\newcommand{\EftbSeedsKittiAbsrelOneMean}{0.170}
\newcommand{\EftbSeedsKittiAbsrelOneSd}{0.013}
\newcommand{\EftbSeedsKittiAlphaMean}{0.03}
\newcommand{\EftbSeedsKittiAlphaSd}{0.05}
\newcommand{\EftbSeedsNyuAbsrelOneMean}{0.326}
\newcommand{\EftbSeedsNyuAbsrelOneSd}{0.026}
\newcommand{\EftbSeedsNyuAlphaMean}{0.09}
\newcommand{\EftbSeedsNyuAlphaSd}{0.03}
\newcommand{\FullDiodeInImages}{325}
\newcommand{\FullDiodeOutImages}{446}
\newcommand{\FullKittiImages}{1000}
\newcommand{\FullNyuImages}{654}
\newcommand{\GeocalibDiodeInBeta}{0.35}
\newcommand{\GeocalibDiodeOutBeta}{0.47}
\newcommand{\GeocalibKittiBeta}{0.72}
\newcommand{\GeocalibNyuBeta}{0.80}

\newcommand{\GeocalibUnshrinkFocalThree}{0.453}
\newcommand{\GeocalibUnshrinkFocalThreecorr}{0.306}

\newcommand{\MetricThreedLAlphaSmallShared}{\ensuremath{-0.04}}

\newcommand{\MetricThreedSAlpha}{\ensuremath{-0.07}}

\newcommand{\MetricThreedSAlphaShared}{\ensuremath{-0.07}}

\newcommand{\MetricThreedSDigitalLr}{1.04}

\newcommand{\MetricThreedSJpeg}{0.99}

\newcommand{\MetricThreedSOptical}{1.05}

\newcommand{\MetricThreedSSrrawOpticalAlpha}{\ensuremath{-0.04}}

\newcommand{\MetricThreedSStaleKAlpha}{0.93}

\newcommand{\MetricThreedSStaleKZoomTwoFactor}{0.53}

\newcommand{\MetricThreedSUncAlphaLarge}{\ensuremath{-0.07}}

\newcommand{\MetricThreedSUncAlphaSmall}{\ensuremath{-0.08}}

\newcommand{\MetricThreedSZoomOfJpeg}{0.98}

\newcommand{\MetricThreedSZoomTwoFactor}{1.05}

\newcommand{\MogeTwoBAlphaShared}{0.37}

\newcommand{\MogeTwoLAlphaShared}{0.35}

\newcommand{\MogeTwoLBetaShared}{0.55}

\newcommand{\MogeTwoSAlpha}{0.36}

\newcommand{\MogeTwoSAlphaLarge}{0.37}

\newcommand{\MogeTwoSAlphaPlusBeta}{0.95}
\newcommand{\MogeTwoSAlphaShared}{0.37}
\newcommand{\MogeTwoSAlphaSmall}{0.27}
\newcommand{\MogeTwoSAlphaSmallShared}{0.33}

\newcommand{\MogeTwoSBeta}{0.56}

\newcommand{\MogeTwoSCamGivenAlpha}{\ensuremath{-0.08}}
\newcommand{\MogeTwoSCamGivenScale}{0.146}

\newcommand{\MogeTwoSCamOwnAlpha}{0.36}
\newcommand{\MogeTwoSCamOwnScale}{0.214}

\newcommand{\MogeTwoSCueMagnify}{0.75}
\newcommand{\MogeTwoSCueMagnifyBelowOne}{4}

\newcommand{\MogeTwoSCueShrinkFocal}{0.75}

\newcommand{\MogeTwoSDigitalLr}{0.82}

\newcommand{\MogeTwoSFixDiodeOutTarget}{0.379}

\newcommand{\MogeTwoSFixDiodeOutZoomed}{3.037}

\newcommand{\MogeTwoSFullDiodeInAlpha}{0.42}

\newcommand{\MogeTwoSJpegAfterZoom}{1.05}

\newcommand{\MogeTwoSOptical}{0.77}

\newcommand{\MogeTwoSPpGuessTwo}{0.496}
\newcommand{\MogeTwoSPpQuadTwo}{0.496}

\newcommand{\MogeTwoSRegionDiff}{0.03}
\newcommand{\MogeTwoSRegionObject}{0.41}
\newcommand{\MogeTwoSRegionObjectLarger}{69}
\newcommand{\MogeTwoSRegionP}{2.9e-12}

\newcommand{\MogeTwoSRegionSurface}{0.39}

\newcommand{\MogeTwoSSpreadRho}{0.02}
\newcommand{\MogeTwoSSrrawDigitalAlpha}{0.26}

\newcommand{\MogeTwoSSrrawOpticalAlpha}{0.38}

\newcommand{\MogeTwoSUncBetaLarge}{0.55}

\newcommand{\MogeTwoSUncBetaSmall}{0.65}

\newcommand{\MogeTwoSZoomOfJpeg}{1.18}

\newcommand{\MogeTwoSZoomOfJpegKitti}{1.28}

\newcommand{\ProxyDiodeIn}{0.88}
\newcommand{\ProxyDiodeOut}{\ensuremath{-0.38}}
\newcommand{\ProxyKitti}{0.98}
\newcommand{\ProxyNyu}{0.95}
\newcommand{\RiskDiodeIn}{0.78}
\newcommand{\RiskDiodeOut}{0.92}
\newcommand{\RiskKitti}{0.92}
\newcommand{\RiskNyu}{0.81}

\newcommand{\SimBimodalRhoSixFiveZerr}{0.54}
\newcommand{\SimBimodalRhoSixFiveZerrc}{0.48}

\newcommand{\UnidepthTwoLAlphaShared}{0.23}

\newcommand{\UnidepthTwoLBetaShared}{0.68}

\newcommand{\UnidepthTwoSAlphaLarge}{0.36}

\newcommand{\UnidepthTwoSAlphaPlusBeta}{0.87}
\newcommand{\UnidepthTwoSAlphaShared}{0.35}
\newcommand{\UnidepthTwoSAlphaSmall}{0.29}
\newcommand{\UnidepthTwoSAlphaSmallShared}{0.18}

\newcommand{\UnidepthTwoSBetaShared}{0.39}

\newcommand{\UnidepthTwoSBlur}{4.1}
\newcommand{\UnidepthTwoSCamGivenAlpha}{0.13}
\newcommand{\UnidepthTwoSCamGivenScale}{0.104}

\newcommand{\UnidepthTwoSCamOwnAlpha}{0.34}
\newcommand{\UnidepthTwoSCamOwnScale}{0.138}

\newcommand{\UnidepthTwoSCueContext}{0.75}
\newcommand{\UnidepthTwoSCueContextBelowOne}{4}

\newcommand{\UnidepthTwoSCueMagnify}{1.07}

\newcommand{\UnidepthTwoSCueShrinkFocal}{0.64}

\newcommand{\UnidepthTwoSFocalThree}{1.63}

\newcommand{\UnidepthTwoSPpQuadTwo}{0.494}

\newcommand{\UnidepthTwoSSrrawDigitalAlpha}{0.39}

\newcommand{\UnidepthTwoSSrrawOpticalAlpha}{0.62}

\newcommand{\UnidepthTwoSUnshrinkBeta}{0.50}

\newcommand{\UnidepthTwoSZoomOfJpeg}{1.16}

\newcommand{\UnidepthTwoSZoomOfJpegKitti}{1.44}

\newcommand{\UnikThreedLAlphaShared}{0.17}

\newcommand{\UnikThreedLBetaShared}{0.82}

\newcommand{\VggtBetaSmallShared}{0.73}

\newcommand{\ZoedepthAlpha}{0.61}

\newcommand{\ZoedepthAlphaSmallShared}{0.54}

\newcommand{\ZoedepthCueContext}{0.95}

\newcommand{\ZoedepthCueMagnify}{0.66}

\newcommand{\ZoedepthFullDiodeInAlpha}{0.43}

\newcommand{\cueContextMax}{1.24}
\newcommand{\cueContextMin}{0.77}
\newcommand{\cueMagnifyBelowOne}{8}
\newcommand{\cueMagnifyMax}{0.72}
\newcommand{\cueMagnifyMin}{0.49}
\newcommand{\cuePixPairs}{8}
\newcommand{\depthproGeocalibAlpha}{0.35}
\newcommand{\depthproGeocalibImages}{800}
\newcommand{\depthproGeocalibScale}{0.226}
\newcommand{\eftbSeeds}{3}
\newcommand{\eftfEpochs}{40}

\newcommand{\eftfValZoomAfter}{0.074}
\newcommand{\eftfValZoomBefore}{0.187}
\newcommand{\fitRmsMedian}{0.061}
\newcommand{\fixedShapeThree}{0.075}
\newcommand{\fixedShiftThree}{1.00}
\newcommand{\fullMaxShift}{0.04}
\newcommand{\fullMedianShift}{0.01}
\newcommand{\geocalibBeta}{0.60}
\newcommand{\geocalibFocalErr}{1.29}
\newcommand{\geocalibImages}{800}
\newcommand{\nImages}{800}
\newcommand{\nLadderShared}{151}
\newcommand{\nScenes}{158}
\newcommand{\nSmallShared}{78}
\newcommand{\proxyPairs}{32}
\newcommand{\proxyRhoDeltaThreeTwoWay}{0.67}
\newcommand{\proxyRhoDeltaThreeTwoWayHi}{0.97}
\newcommand{\proxyRhoDeltaThreeTwoWayLo}{0.08}
\newcommand{\proxyRhoOne}{\ensuremath{-0.09}}
\newcommand{\proxyRhoThree}{0.55}
\newcommand{\proxyRhoThreeHi}{0.86}
\newcommand{\proxyRhoThreeLo}{0.18}

\newcommand{\proxyRhoThreeTwoWayHi}{0.97}
\newcommand{\proxyRhoThreeTwoWayLo}{\ensuremath{-0.23}}
\newcommand{\riskAurc}{0.86}
\newcommand{\riskBelowOne}{26}
\newcommand{\riskPairs}{28}
\newcommand{\srrawSequences}{50}
\newcommand{\srrawShots}{150}
\newcommand{\staleDiff}{1.000}
\newcommand{\staleDiffMean}{0.999}
\newcommand{\staleImages}{800}
\newcommand{\udDiodeInAbsrelOne}{0.25}
\newcommand{\udDiodeOutAbsrelOne}{2.75}
\newcommand{\udDiodeOutOverTwo}{38}
\newcommand{\udDiodeOutScale}{1.44}

\newcommand{\udNyuAbsrelOne}{0.11}

\title{Zoom In, Scale Out: Metric Depth Models Mistake Digital Zoom for Moving Closer}

\author{
Rabeya Noor\\
Department of CSE\\
Daffodil International University\\
\href{mailto:rabeyanoor.dev@gmail.com}{\tt\small rabeyanoor.dev@gmail.com}
\and
Khandoker Nosiba Arifin\\
Department of CSE\\
Daffodil International University\\
\href{mailto:nosibaarifin.cse@diu.edu.bd}{\tt\small nosibaarifin.cse@diu.edu.bd}
}

\begin{document}
\maketitle
\begin{abstract}
Many new AI models estimate how far away things are, in metres, from a single photo. To do this they must also
work out what camera took the photo. We check whether they really do with a simple test: digital zoom, that is,
cropping a photo and enlarging the crop back to full size. Zooming does not move anything in the scene, so a
correct model should give the same distance for every point that is still in the picture. The test is free, exact
and needs no ground-truth labels. We describe a model's reaction with two numbers: how much its depth changes
with the zoom factor $s$ (written $s^{-\alpha}$; ideal $\alpha=0$) and how much its estimate of the camera's focal
length grows (written $s^{\beta}$; ideal $\beta=1$). On \nImages{} images from KITTI, NYUv2 and DIODE, the widely
used Depth Anything V2 has $\alpha=\DavTwoSAlpha$: it treats a zoomed photo almost exactly as if the camera had
moved closer, because it judges distance mainly from how large things look. Models that also predict the camera
remove only part of this error ($\alpha\approx\DepthproAlpha$--\MogeTwoSAlpha), because their focal-length
estimate grows too little. When we give Depth Pro and DA3-Nested the true camera, their depth stays correct
($\alpha\approx0$), so the error comes from their camera estimate, not from their depth estimate. We explain why:
a zoomed photo is, pixel for pixel, identical to a normal photo of a scene squashed in depth, so an object's size
alone cannot tell the two apart, and a camera estimator trained by regression plays safe and stays close to the
cameras it saw in training. Undoing this ``playing safe'' needs no labels and no training, and it removes Depth
Pro's error ($\alpha=\DepthproUnshrinkAlphacorr$) at almost no cost on normal photos. Control tests, including a
real optical zoom lens, rule out image blur and resizing artefacts as the cause. Finally, fine-tuning on unlabeled
photos so that the model agrees with itself across zooms lowers $\alpha$ to \DavTwoSEftbKittiAlpha{} on KITTI and
\DavTwoSEftbNyuAlpha{} on NYUv2, once we stop the model from learning a ``blurry means zoomed'' shortcut; we also
report what this costs on normal and unseen photos.
Code: \url{https://github.com/rabeyanoor/zoom-equivariance}.
\end{abstract}

\section{Introduction}
\label{sec:intro}

\begin{figure*}[t]
  \centering
  \IfFileExists{fig/teaser.pdf}{\includegraphics[width=\linewidth]{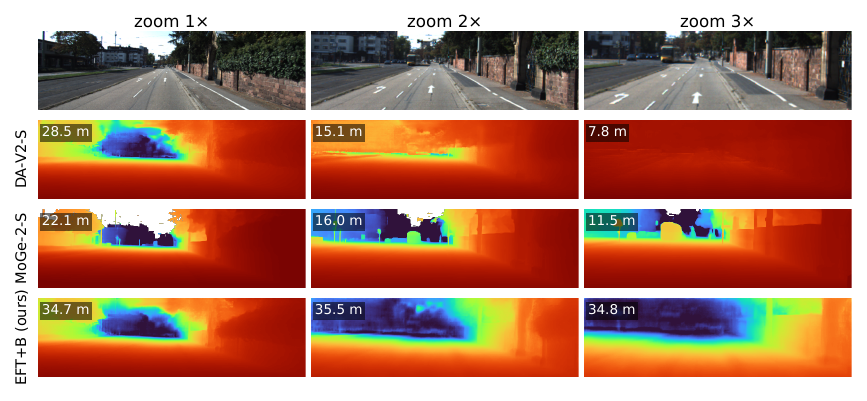}}{\fbox{teaser pending}}
  \caption{A KITTI image at $1\times$, $2\times$, $3\times$ digital zoom, one colour scale (metres); the number is
  the median depth inside the $3\times$ crop, the same points in every column. Depth Anything V2 moves the
  scene closer (28.5\,m to 7.8\,m), MoGe-2 less. Our label-free EFT+B (392-px input) keeps it (34.7, 35.5,
  34.8\,m; frozen at that input: 33.6 to 11.8\,m) at a wrong scale (LiDAR: 21.3\,m). NYUv2: supplement.}
  \label{fig:teaser}
\end{figure*}

\paragraph{The problem in plain words.}
Models that tell how far away things are, in metres, from one photo (\emph{metric depth}) have improved
fast~\cite{bochkovskii2025depthpro,piccinelli2025unidepthv2,wang2025moge2,hu2024metric3dv2,yang2024dav2}.
They now run on phones, robots and cars, often with cameras they never saw in training. But a single photo
hides an old problem: a close scene taken with a wide-angle lens and a far scene taken with a telephoto lens
can look the same~\cite{he2018focal}. To answer in metres, a model has to work out the camera too. Some models
are given the camera's settings (its \emph{intrinsics}, mainly the focal length). Some predict them. Many simply
guess metres from the pixels, for example from the usual size of cars or doors. Benchmark scores cannot show
which of these really happens, because test sets come from only a few cameras.

\paragraph{Our test: digital zoom.}
We use a test that needs no labels and no new data. Crop a photo and enlarge the crop back to full size. The
result is \emph{exactly} the photo a camera with a longer focal length would have taken from the same spot.
Nothing in the scene has moved, so every point still in view is at the same distance as before; only the
focal length has grown, by the zoom factor $s$ (\cref{sec:law}). A model that understands the camera should
therefore keep its depth the same and multiply its focal-length estimate by $s$. This matters in practice:
phones zoom digitally all the time, and so do gallery apps, screenshots and object-detection crops.
Earlier work showed that depth maps change under cropping~\cite{zhong2023equivariance} and that models which are
given the camera fail when a crop leaves the camera settings out of date~\cite{kim2026mismatch}. What is still
missing is a test of today's metric models in which the cropped photo's camera is updated correctly.

We ask four questions.
\begin{itemize}\setlength{\itemsep}{0pt}
  \item[\textbf{Q1}] Do current metric depth models keep the same depth when a photo is digitally zoomed, and
        do models that predict the focal length multiply it by $s$?
  \item[\textbf{Q2}] Do models that are given, predict, or ignore the camera fail in different ways, and can
        geometry explain why?
  \item[\textbf{Q3}] Which part of a zoom causes the failure: losing the area around the crop, or everything
        left in the crop looking larger?
  \item[\textbf{Q4}] Can the zoom test itself be used to train a model on unlabeled photos, and what does that
        cost?
\end{itemize}

\paragraph{What we find.}
\emph{Q1.} No tested model that has to guess or ignores the camera passes the test. Depth Anything V2 makes the
scene look closer almost in proportion to the zoom ($\alpha=\DavTwoSAlpha$, where $\alpha=1$ would mean
``exactly like walking closer'' and $\alpha=0$ means correct; \cref{fig:teaser}). Models with a camera head do
better, but their focal-length estimate grows far less than $s$ (\cref{fig:curves}).
\emph{Q2.} All failures follow one pattern. A zoomed photo is, pixel for pixel, the same as a normal photo of the
scene squashed along the viewing direction; we call this the \emph{zoom--squash ambiguity}. A model that judges
distance from size cannot tell the two apart, and lands on the line $\alpha+\beta=1$ (with $\beta$ the growth of
its focal estimate), which is where Depth Pro sits image by image. Giving the network the true camera shows where
the error lives: Depth Pro and DA3-Nested then keep their depth ($\alpha=\DepthproCamGivenAlpha$ and
\DaThreeNestedCamGivenAlpha), so the whole error comes from their focal-length estimate (\cref{tab:camera}). The
same ambiguity explains why every focal estimator we test, including a dedicated calibration network, grows too
little: trained by regression, it stays close to the cameras it saw in training. Undoing this, without labels,
makes Depth Pro pass the test.
\emph{Q3.} For models that use only pixels, the failure comes from the content looking larger, not from the lost
surroundings, on every dataset.
\emph{Q4.} Fine-tuning a model so that it agrees with itself across zooms, in the spirit
of~\cite{zhong2023equivariance}, first teaches it a shortcut, ``blurry means zoomed''. Adding a term that makes the
model ignore blur removes the shortcut and brings $\alpha$ close to zero on the training datasets, at a cost on
other photos that we report in full.

\paragraph{Contributions.}
\begin{enumerate}\setlength{\itemsep}{0pt}
  \item A label-free \textbf{digital-zoom test} for metric depth, with exact geometry and two simple numbers,
        one for depth ($\alpha$) and one for focal length ($\beta$). It comes with control tests for resizing,
        loss of detail, image compression, crop position, and optical against digital zoom, including a real zoom
        lens (\cref{sec:audit}). We also check what the score predicts: how much a model degrades under zoom,
        not its accuracy in general.
  \item The classical focal--depth ambiguity written as an exact, testable \textbf{zoom--squash} statement, and the
        $\alpha+\beta=1$ line it predicts for the main failure. We check it on three groups of models and a range
        of models up to Depth Anything~3 and VGGT (\cref{sec:law,sec:results}).
  \item Two \textbf{experiments that locate the failure}. Giving the same network the true camera shows that the
        focal-length head, not the depth network, causes the error of Depth Pro and DA3-Nested, and that their
        depth fits their own focal estimate better than the true one, whereas MoGe-2 and UniDepthV2 improve in
        both zoom consistency and accuracy (\cref{sec:camint,tab:camera}). An account of why focal heads ``play
        safe'' ($\beta$ measures how reliable the camera cue is) gives \emph{FocalUnshrink}, a correction that needs
        no training and no labels. Splitting a zoom into lost surroundings and enlargement shows that enlargement
        drives the pixels-only models (\cref{sec:cues,tab:cues}).
  \item A careful test of \textbf{label-free correction}: fine-tuning for self-consistency, the blur shortcut it
        learns, EFT+B that removes it, a scale-only variant (EFT-S) that trains on a CPU from a cache, a
        training-free fix when the zoom factor is known, and the accuracy cost on unseen data
        (\cref{sec:eft,sec:eftresults}).
\end{enumerate}

\section{Related work}
\label{sec:related}

\paragraph{Metric depth and the camera.}
Early work gave the focal length to the network~\cite{he2018focal} or made convolutions
camera-aware~\cite{facil2019camconvs}; FS-Depth~\cite{wei2023fsdepth} augments focal lengths by cropping, with
depth labels. Today's models fall into three groups. Some \emph{take the camera}: Metric3D~\cite{yin2023metric3d,hu2024metric3dv2}
maps every image to a canonical camera and needs $K$ at test time, and Depth Any Camera~\cite{guo2025dac}
and ZeroDepth~\cite{guizilini2023zerodepth} are also conditioned on it. Some \emph{predict the camera}:
Depth Pro~\cite{bochkovskii2025depthpro} estimates the field of view,
UniDepth~\cite{piccinelli2024unidepth,piccinelli2025unidepthv2} predicts a ray per pixel, and
MoGe-2~\cite{wang2025moge2} returns a metric point map with intrinsics. The rest use \emph{pixels only},
such as the metric heads of Depth Anything~\cite{yang2024da,yang2024dav2} and ZoeDepth~\cite{bhat2023zoedepth}.
Training on many cameras~\cite{ma2026metricanything} is meant to make camera inference robust; our audit
tests whether it is.

\paragraph{Depth cues.}
DepthCues~\cite{danier2025depthcues} probes which monocular cues large vision models pick up, among them
the known size of objects. Our exponent $\alpha$ measures one cue in particular, apparent size, and our
cue intervention (\cref{sec:cues}) changes it directly while keeping everything else fixed.

\paragraph{Single-image calibration.}
Classical methods calibrate from vanishing points and right angles~\cite{hartley2004mvg}; learned ones regress
the intrinsics~\cite{bogdan2018deepcalib}, predict crop- and resize-invariant incidence fields
(WildCamera~\cite{zhu2023wildcamera}) or local perspective fields~\cite{jin2023perspectivefields}, or combine a
network with geometric optimisation (GeoCalib~\cite{veicht2024geocalib}). These are the cues the zoom--squash
ambiguity says a model must use. InFlux~\cite{liang2025influx} benchmarks video
cameras whose intrinsics change over time, for example by zooming, and shows that most calibration methods
struggle with it. We ask instead whether depth models respond correctly to the one camera change whose effect
is known exactly.

\paragraph{Crops and equivariance.}
Zhong \etal~\cite{zhong2023equivariance} are the closest to us. They show that supervised depth and normal
predictors are not equivariant to crop-and-resize and fix it with test-time averaging and a self-consistency
loss on unlabeled images; the output is treated as a map that should simply be cropped, and the camera does
not appear. UniDepth trains a geometric invariance loss that aligns camera-conditioned depth features across
rescaled and cropped views~\cite{piccinelli2024unidepth}. We instead measure how released models change their
metric depth and focal under the exact camera change. For \emph{metric} models a crop is a zoom, the focal length must
grow by $s$, and a model that reads apparent size breaks the rule in a predictable way. Our fine-tuning
(\cref{sec:eft}) follows the self-consistency idea of~\cite{zhong2023equivariance} with a frozen teacher; we do
not claim the loss as new and compare against their stop-gradient target.
Kim and Lee~\cite{kim2026mismatch} study camera-conditioned models whose pipeline passes stale intrinsics after a
crop, resize or padding, and propose a mismatch-aware camera module; our stale-$K$ run is one case of this
setting, included only to put such models on the same scale.
Crops also matter elsewhere: resize-and-crop augmentation breaks plane geometry in self-supervised
depth~\cite{wang2023planedepth}, crops cause a perspective ambiguity for object-level 3D~\cite{prakash2024kpe} and
human meshes~\cite{nie2024multicrop}, and NERO~\cite{zhao2023nero} visualises broken equivariance. Robustness to
camera roll~\cite{tang2026roll} leaves the intrinsics unchanged; a zoom does not.

\paragraph{Label-free evaluation and adaptation.}
Self-Geometry~\cite{youn2026selfgeometry} adapts 3D foundation models at test time with multi-view
constraints, Re-Depth Anything~\cite{bhattarai2025redepth} refines depth by self-supervised re-lighting, and
test-time adaptation has been used for depth completion~\cite{park2024ttadepth}. A model can be consistently wrong, so we always report the
label-free exponents next to errors against ground truth.

\section{Digital zoom as an exact camera transform}
\label{sec:law}

\paragraph{The law.}
Let an image $I$ of size $W\times H$ be taken by a pinhole camera with intrinsics
$K=\begin{psmallmatrix} f_x & 0 & c_x\\ 0 & f_y & c_y\\ 0&0&1\end{psmallmatrix}$.
Cropping the box $b=(x_0,y_0,w,h)$ and resizing it back to $W\times H$ gives an image
$\mathcal{Z}_b I$ that is \emph{exactly} the image of the same scene taken with
\begin{equation}
  f'_x = s_x f_x,\quad c'_x = s_x\,(c_x - x_0 + \tfrac12) - \tfrac12,\qquad s_x = W/w,
  \label{eq:law}
\end{equation}
and likewise for $y$ (the half pixels come from pixel centres).
Every pixel that survives the crop sees the same 3D point, so its metric depth is unchanged.
For a centred crop with $s_x=s_y=s$ we call $s$ the \emph{zoom factor}.
\cref{eq:law} needs no assumption about the scene, the model or the data;
our implementation reproduces it to below $10^{-3}$\,px (supplement).

\paragraph{What a correct model must do.}
Let a model map an image to a depth map $\hat D(I)$ and, optionally, a focal estimate
$\hat f(I)$. Zoom equivariance asks
\begin{equation}
  \hat D(\mathcal{Z}_b I)\big(\mathcal{Z}_b(p)\big) = \hat D(I)(p), \qquad
  \hat f(\mathcal{Z}_b I) = s\,\hat f(I)
  \label{eq:equiv}
\end{equation}
for all pixels $p$ inside $b$. Neither side involves ground truth.

\paragraph{Two exponents.}
Real models do not satisfy \cref{eq:equiv}; over $s\in[1,3]$ we summarise their behaviour by two power laws,
\begin{equation}
  \frac{\hat D(\mathcal{Z}_b I)}{\hat D(I)} \approx s^{-\alpha}, \qquad
  \frac{\hat f(\mathcal{Z}_b I)}{\hat f(I)} \approx s^{\beta},
  \label{eq:exponents}
\end{equation}
fitted per image by least squares in log-log space through the origin
(the depth ratio is the median over the pixels in $b$).
An equivariant model has $\alpha=0$ and $\beta=1$. The fit is an approximation: the median residual is
\fitRmsMedian{} in log depth, and $\alpha$ grows with the zoom (Depth Pro \DepthproAlphaSmall{} for $s\le1.5$,
\DepthproAlphaLarge{} for $s\ge2$) because $\beta$ falls (\DepthproUncBetaSmall{} to \DepthproUncBetaLarge) while
$\alpha+\beta$ stays near 1 (supplement).
A model that ignores the zoom in its camera estimate has $\beta=0$.

\paragraph{Three invariants.}
Unprojecting the prediction with the model's own focal gives a point cloud.
Under a zoom, depth scales by $s^{-\alpha}$ and the lateral coordinate
$X=(u-c)\,\hat D/\hat f$ by $s\cdot s^{-\alpha}/s^{\beta}=s^{1-\alpha-\beta}$. Hence
\begin{itemize}
  \item $\alpha=0$: depth is preserved;
  \item $\beta=1$: the shape of the point cloud (the ratio $X/Z$) is preserved;
  \item $\alpha+\beta=1$: lateral coordinates are preserved (not depth, not shape).
\end{itemize}
The third relation is what a model obtains if it reasons from apparent size.
An object of metric width $L$ that spans $\ell$ pixels is placed at
$\hat D=\hat f L/\ell$. A zoom multiplies $\ell$ by $s$ and $\hat f$ by $s^{\beta}$, so
$\hat D$ scales by $s^{\beta-1}$, that is $\alpha=1-\beta$.
A pixels-only model ($\hat f$ fixed, implicitly) that reasons from size has $\alpha=1$:
twice as large means twice as close.

\paragraph{The zoom--squash ambiguity.}
Why would a model reason from size at all, when the zoom is in plain sight?
Because a zoom is invisible to size cues.
A point $(X,Y,Z)$ seen with focal $sf$ projects to $u = sf\,X/Z + c = f\,X/(Z/s) + c$.
\begin{quote}
\emph{A zoomed photo of a scene is pixel-for-pixel the unzoomed photo of the same scene
squashed along the viewing axis by $s$.}
\end{quote}
This is the classical focal--depth ambiguity of a single view~\cite{he2018focal}, written as an exact, testable
statement about a zoom; it is not a new ambiguity.
The squashed scene has the same lateral sizes, so every object has a plausible width; only its
extent \emph{in depth} is wrong (a 4.5\,m car becomes 2.25\,m long at $s=2$), and so are the
angles between surfaces. Telling the two apart needs priors on 3D shape (aspect ratios in
depth, right angles, the rate at which parallel lines converge), not priors on size.
One consequence is a hard limit: a fronto-parallel plane squashed along the viewing axis is still a fronto-parallel plane, so
\emph{no} estimator can be equivariant on such scenes without being told the camera.
The ambiguity also explains why the third invariant is the natural failure mode: believing the
squashed scene means $\alpha=1$ with $\beta=0$, a point on the line $\alpha+\beta=1$.
Note that the line only guarantees lateral consistency; along it, depth and shape are wrong
unless $\beta=1$.

\paragraph{Why a focal head under-scales.}
A simple model shows why a regression-trained focal head has $\beta<1$. Suppose the training cameras
spread as $\log f\sim\mathcal N(\mu,\tau^2)$ and the image carries evidence $x=\log f+\epsilon$,
$\epsilon\sim\mathcal N(0,\sigma^2)$. The squared-error optimum is the posterior mean $\mu+\rho\,(x-\mu)$ with
$\rho=\tau^2/(\tau^2+\sigma^2)$. A zoom adds $\log s$ to $x$, so $\beta=\rho$: the focal exponent is the
\emph{reliability} of the camera cue, and it equals 1 only for a perfect cue. Inverting the shrinkage,
$\log f_c=\mu+(\log\hat f-\mu)/\beta$, restores $\beta=1$ but scales the noise by $1/\beta$, so equivariance
and accuracy trade off more steeply the less reliable the cue. Both $\beta$ and $\mu$ can be measured without
labels (\cref{sec:results}).

\paragraph{Models that take the camera as input.}
Metric3D~\cite{yin2023metric3d,hu2024metric3dv2} receives $K$ and predicts depth in a canonical camera, so with
the correct $K'$ it is equivariant by design. When the zoom happens after capture (a gallery crop, a screenshot)
the metadata may still describe the original camera~\cite{kim2026mismatch}; with this \emph{stale} $K$ only the
final scale $f/1000$ changes, by $s$, so $\alpha_{\text{stale}}=\alpha_{\text{true}}+1$ up to output clipping. Our runs confirm this (mean difference over \staleImages{} images
\staleDiffMean), so a camera-conditioned model with stale metadata behaves like a pixels-only model that
reads apparent size.

\section{The zoom audit}
\label{sec:audit}

\paragraph{Models.}
We test seven models in three groups (\cref{sec:related}):
\emph{pixels only} --- Depth Anything V2 metric Small and Base~\cite{yang2024dav2} (indoor
or outdoor head, matching the dataset) and ZoeDepth NYU+KITTI~\cite{bhat2023zoedepth};
\emph{predicts camera} --- Depth Pro~\cite{bochkovskii2025depthpro},
MoGe-2 ViT-S~\cite{wang2025moge2} and UniDepthV2 ViT-S~\cite{piccinelli2025unidepthv2};
\emph{takes camera} --- Metric3D v2 ViT-S~\cite{hu2024metric3dv2}, run once with the correct
zoomed $K'$ and once with the stale $K$.
All use the official weights and the authors' preprocessing (Hugging Face
\texttt{transformers} or the official repositories).

\paragraph{Data.}
KITTI depth val selection~\cite{geiger2013kitti,uhrig2017sparsity} (outdoor, LiDAR),
the 654 official NYUv2 test images~\cite{silberman2012nyu} (indoor, Kinect, Eigen crop),
and the DIODE validation set~\cite{vasiljevic2019diode} split into indoor and outdoor scans
(laser). From each of the four sets we draw a fixed random subset of 200 images (seed 0),
identical for all models. Ground-truth intrinsics come with all three datasets.

\paragraph{Zooms and controls.}
We zoom by $s\in\{1.25, 1.5, 2, 3\}$ about the image centre (bicubic, antialiased, as in a
phone's digital zoom). The zoom factor used in all fits is the exact one, $f'_x/f_x$ from
\cref{eq:law}, which absorbs the rounding of the crop box. The prediction on the zoomed image
is resampled onto the crop's own pixel grid by area averaging and compared, pixel by pixel,
with the prediction on the original image (label-free) and with ground truth (for accuracy).
To separate the camera change from the loss of detail, the \emph{blur control} down-samples
the image by $s=2$ and up-samples it again: same detail loss as a $2\times$ zoom, same field of
view. A subset of 50 images per set from KITTI and NYUv2 is also zoomed about the four
quadrant centres, where the principal point moves out of the image centre.

\paragraph{Optical versus digital zoom.}
A digital zoom changes the camera \emph{and} loses detail. We treat a $2\times$ down-sampled copy of each image
as the camera. Its \emph{optical} $2\times$ zoom is the same view cut from the full-resolution image, with real
detail. Its \emph{digital} $2\times$ zoom is a crop of the down-sampled image, up-sampled. Both have the same
intrinsics. We also test bilinear resampling and JPEG quality 75.

\paragraph{Splitting a zoom into its causes.}
\label{sec:cues}
A digital zoom removes the context around the crop and makes everything in it $s$ times larger. To separate
the two, \emph{paste} keeps the $2\times$ crop at its own place and size on a mean-colour canvas (original $K$,
only the context is gone), and \emph{shrink} places the whole image $2\times$ smaller on a plain canvas, with its
own intrinsics. On the same pixels we compare paste with original (\emph{context}), zoom with paste
(\emph{magnification}), zoom with original, and shrink with original; apparent-size reasoning gives 1, 0.5, 0.5
and 2, a camera-aware model 1 throughout. The canvas is unusual, so these are interventions on the cue, not
natural images. Samples: 100 images per set (50 for Depth Pro and DA3-Metric-L).

\paragraph{Giving the camera back.}
\label{sec:camint}
To locate the failure of a camera-predicting model, we give the same network the true zoomed camera $K'$ of
\cref{eq:law}. For Depth Pro and DA3-Nested, whose metres are their own focal estimate times a camera-free
output, this is an exact rescaling of the audit's predictions. MoGe-2 takes the true horizontal field of view in
place of the one it recovers from its point map, and UniDepthV2 conditions its decoder on the rays of $K'$; we
rerun both on the same 200 images per set as their plain runs.

\paragraph{Metrics.}
\textbf{Label-free:} the exponents $\alpha$ and $\beta$ of \cref{eq:exponents}, and the depth
inconsistency at $2\times$, $\mathrm{DIE}_2=|\operatorname{median}_p \log
\hat D_2/\hat D_1|$. \textbf{With labels:} AbsRel and $\delta_1$ on the pixels of the crop
that have ground truth (capped at 80\,m for KITTI, 10\,m for NYUv2, 50\,m and 150\,m for
DIODE indoor and outdoor). We report medians over images with 95\% bootstrap intervals. Frames of one KITTI drive, one
DIODE scene or one NYUv2 scene are not independent, so the intervals in the main tables resample
whole scenes (cluster bootstrap); paired comparisons use Wilcoxon signed-rank tests with Holm
correction. We also report a shape error, the median absolute deviation of
$\log \hat D_s/\hat D_1$ from its median, which is zero when a zoom changes depth by a single global
factor.

\section{Results: the zoom audit}
\label{sec:results}

\newcommand{\maybefig}[1]{\IfFileExists{fig/#1.pdf}{\includegraphics[width=\linewidth]{fig/#1.pdf}}{\fbox{\small figure \detokenize{#1} pending}}}

\begin{table*}[t]\centering\small
\caption{Zoom audit, pooled over KITTI, NYUv2, DIODE-indoor and DIODE-outdoor (\nImages{} images per model, \nScenes{} scenes). $\alpha$: depth exponent (ideal 0; apparent size only: 1), median over images with a 95\% interval from a scene-level bootstrap. $\beta$: focal exponent (ideal 1). $\alpha+\beta$: median per image (1 = lateral sizes kept, \cref{sec:law}). Blur: depth change under the blur control. AbsRel at $1\times$ on the full image, and at $3\times$ on the crop for the unzoomed prediction (``same px'') and the zoomed one. The high AbsRel of UniDepthV2-S at $1\times$ is DIODE-outdoor alone (\udDiodeOutAbsrelOne; other sets \udNyuAbsrelOne--\udDiodeInAbsrelOne), a scale over-estimate (\cref{sec:results}).}
\label{tab:audit}
\setlength{\tabcolsep}{4pt}
\begin{tabular}{llccccccc}\toprule
& & & & & & \multicolumn{3}{c}{AbsRel} \\ \cmidrule(lr){7-9}
Model & Group & $\alpha$ & $\beta$ & $\alpha+\beta$ & Blur & $1\times$ & $3\times$ same px & $3\times$ zoomed \\ \midrule
DA-V2-S & pixels only & 0.95 [0.83, 1.12] & -- & -- & 1.3\% & 0.317 & 0.336 & 0.652 \\
DA-V2-B & pixels only & 0.85 [0.76, 0.94] & -- & -- & 0.8\% & 0.313 & 0.335 & 0.599 \\
ZoeDepth & pixels only & 0.61 [0.53, 0.75] & -- & -- & 0.7\% & 0.310 & 0.302 & 0.539 \\
Depth Pro & predicts camera & 0.31 [0.20, 0.41] & 0.65 & 0.97 & 2.8\% & 0.310 & 0.293 & 0.459 \\
MoGe-2-S & predicts camera & 0.36 [0.30, 0.40] & 0.56 & 0.95 & 3.1\% & 0.274 & 0.277 & 0.465 \\
UniDepthV2-S & predicts camera & 0.34 [0.23, 0.51] & 0.45 & 0.87 & 4.1\% & 0.824 & 0.785 & 0.792 \\
Metric3D v2-S & takes camera (true K) & \ensuremath{-0.07} [\ensuremath{-0.11}, \ensuremath{-0.05}] & -- & -- & 1.6\% & 0.142 & 0.148 & 0.358 \\
Metric3D v2-S (stale K) & takes camera (stale K) & 0.93 [0.89, 0.95] & -- & -- & 1.6\% & 0.142 & 0.148 & 0.628 \\
\bottomrule\end{tabular}\end{table*}

\begin{figure}[t]
  \centering
  \maybefig{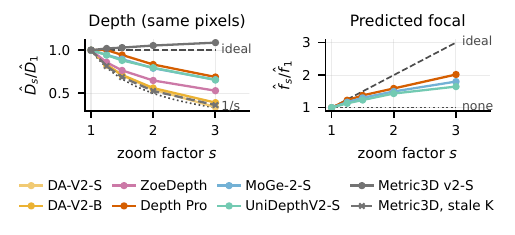}
  \caption{Depth (left) and predicted focal (right) of the same pixels as a function of the zoom factor,
  median over images. Dashed: the law; dotted: apparent size only.}
  \label{fig:curves}
\end{figure}

\paragraph{Models that infer or ignore the camera miss the zoom (Q1).}
\Cref{tab:audit} and \cref{fig:curves} summarise \nImages{} images per model. The pixels-only models
shrink depth almost like $1/s$: $\alpha=\DavTwoSAlpha$ [\DavTwoSAlphaLo, \DavTwoSAlphaHi] for
DA-V2-S, \DavTwoBAlpha{} for DA-V2-B and \ZoedepthAlpha{} for ZoeDepth. A $3\times$ zoom thus moves a
scene to about \DavTwoSRatioThree{} of its predicted distance. Models that predict their camera do
better but are far from equivariant ($\alpha$ between \DepthproAlpha{} and \MogeTwoSAlpha), and their
focal estimate grows by only \UnidepthTwoSFocalThree--\DepthproFocalThree{} at $3\times$ instead of
$3$. Metric3D with the correct $K'$ is close to the law ($\alpha=\MetricThreedSAlpha$, a slight
over-correction), and with the stale $K$ its exponent is $\alpha=\MetricThreedSStaleKAlpha$, larger by
\staleDiff{} per image (median) as derived in \cref{sec:law}. On the same pixels, the AbsRel of DA-V2-S grows from \DavTwoSBaseThree{} unzoomed to \DavTwoSAbsrelThree{}
at $3\times$.

\begin{figure}[t]
  \centering
  \IfFileExists{fig/fig3_alpha_beta_plane.pdf}{\includegraphics[width=0.78\linewidth]{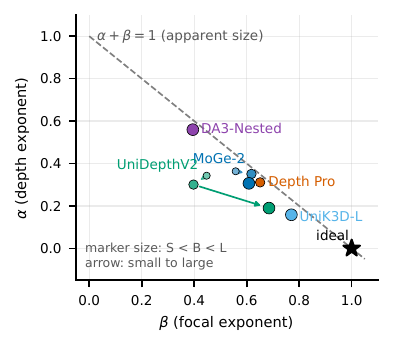}}{\fbox{pending}}
  \caption{Camera-predicting models in the $(\beta,\alpha)$ plane. Dashed: $\alpha+\beta=1$, where
  lateral sizes are preserved (apparent-size reasoning). The ideal model is at $(1,0)$. Arrows join the
  sizes of one family, small to large; each model on its own sample (supplement).}
  \label{fig:abplane}
\end{figure}

\paragraph{The failure follows apparent size (Q2).}
\Cref{fig:abplane} places each camera-predicting model in the $(\beta,\alpha)$ plane. Depth Pro lies on
the line $\alpha+\beta=1$ (median of the per-image sums \DepthproAlphaPlusBeta), MoGe-2 close to it
(\MogeTwoSAlphaPlusBeta) and UniDepthV2 below it (\UnidepthTwoSAlphaPlusBeta). On this line a model keeps
the lateral size of what it sees but not its depth, the reading the zoom--squash ambiguity makes available:
the zoomed image is consistent with a compressed scene in which every object keeps its width. The line is a
geometric signature, not by itself a proof of how a model reasons; for Depth Pro it follows from its
architecture, as the next paragraphs show. The ambiguity explains the direction of the failure, not which images escape it: scenes with a larger
depth range are not easier, and familiar objects such as people and cars follow the zoom at least as
much as walls and floors (supplement).

\paragraph{It is the camera, not the pixels.}
The blur control, which loses as much detail as a $2\times$ zoom without changing the field of view,
moves depth by only \DavTwoSBlur\% for DA-V2-S and at most \UnidepthTwoSBlur\% for any model in
\cref{tab:audit}. A further control (supplement) separates the two effects of a digital zoom directly: an
\emph{optical} $2\times$ zoom with real detail changes the depth of DA-V2-S by a factor of
\DavTwoSOptical, the same zoom done digitally by \DavTwoSDigitalLr, while bilinear resampling and JPEG
compression change it by \DavTwoSBilinear{} and \DavTwoSJpeg.
Real detail does not rescue the model. On a real zoom lens (SR-RAW~\cite{zhang2019zoom}, \srrawSequences{}
tripod sequences, supplement) DA-V2-S gives $\alpha=\DavTwoSSrrawOpticalAlpha$, the same as for the
digital zoom of the widest shot (\DavTwoSSrrawDigitalAlpha). Depth Pro keeps its depth under the real lens
($\alpha=\DepthproSrrawOpticalAlpha$), as on DIODE-outdoor (\DepthproDsDiodeOutAlpha), but not on KITTI, NYUv2 and
DIODE-indoor (\DepthproDsKittiAlpha, \DepthproDsNyuAlpha, \DepthproDsDiodeInAlpha): its failure depends on the scene.

\paragraph{Bigger and newer models.}
The supplement adds larger variants, Depth Anything~3~\cite{lin2025da3}, UniK3D~\cite{piccinelli2025unik3d} and
VGGT~\cite{wang2025vggt}. On the \nLadderShared{} images every model saw, size helps but does not remove the effect
($\alpha$ \DavTwoSAlphaShared, \DavTwoBAlphaShared, \DavTwoLAlphaShared{} for DA-V2-S, B, L), and UniK3D-L comes
closest ($\alpha=\UnikThreedLAlphaShared$). The largest model, DA3-Nested (\DaThreeNestedParams M), shrinks depth
with $\alpha=\DaThreeNestedAlphaSmallShared$ on the \nSmallShared{} images it shares with the main models, near ZoeDepth (\ZoedepthAlphaSmallShared), while DA3-Metric-L and
Metric3D v2-L, given the camera, keep it (\DaThreeMetricLAlphaSmallShared, \MetricThreedLAlphaSmallShared). Whether
the camera enters the model matters more than its size.

\paragraph{Same network, true camera.}
Is the failure in the camera head or in the depth network? Depth Pro and DA3-Nested turn a camera-free
network output into metres by multiplying it with their own focal estimate (\texttt{infer} in the official
Depth Pro code~\cite{bochkovskii2025depthpro}; the metric branch of DA3-Nested~\cite{lin2025da3}). Giving them
the true zoomed focal instead therefore changes each depth map by the exact factor $f^{*}/\hat f$. It removes the failure: $\alpha$ falls from \DepthproCamOwnAlpha{} to \DepthproCamGivenAlpha{} for
Depth Pro and from \DaThreeNestedCamOwnAlpha{} to \DaThreeNestedCamGivenAlpha{} for DA3-Nested
(\cref{tab:camera}). For such a model, $\hat D=\hat f\,C(I)$ with a camera-free $C$, so per image
$\alpha+\beta=1+\alpha_{K}$ exactly, where $\alpha_{K}$ is the exponent with the true focal (it holds on every
image to machine precision). Their camera-free outputs follow the zoom as geometry requires ($\alpha_{K}\approx0$),
the focal head carries the whole error, and this is why they sit on $\alpha+\beta=1$. A calibration-only network, GeoCalib~\cite{veicht2024geocalib}, under-scales the zoomed focal as
well ($\beta=\geocalibBeta$, supplement). The true focal does not make
them more accurate on unzoomed images, however. Their median metric-scale error rises from
\DepthproCamOwnScale{} to \DepthproCamGivenScale{} and from \DaThreeNestedCamOwnScale{} to
\DaThreeNestedCamGivenScale{} (it rises on DIODE for both and is about unchanged on NYUv2; the exception is Depth Pro on KITTI,
\DepthproCamOwnScaleKitti{} to \DepthproCamGivenScaleKitti; supplement). Overall the depth output is better matched to the model's own focal than to the true
one, so a focal estimate is not simply a camera measurement that can be swapped for EXIF. MoGe-2 and
UniDepthV2 also accept the camera as input, and for them the true camera helps on both counts: $\alpha$ falls
from \MogeTwoSCamOwnAlpha{} to \MogeTwoSCamGivenAlpha{} and from \UnidepthTwoSCamOwnAlpha{} to
\UnidepthTwoSCamGivenAlpha, and the unzoomed scale error from \MogeTwoSCamOwnScale{} to \MogeTwoSCamGivenScale{}
and from \UnidepthTwoSCamOwnScale{} to \UnidepthTwoSCamGivenScale. UniDepthV2-S's large error on DIODE-outdoor
(\cref{tab:audit}) is a scale over-estimate (median factor \udDiodeOutScale), which the official inference code
reproduces.
\begin{table}[t]\centering\small
\caption{The same camera-predicting network with its own focal and given the true (zoomed) camera. $\alpha$: depth exponent. Scale: median over images of $|\operatorname{median}_p \log \hat D/D^{*}|$ at $1\times$ (0 = correct metric scale). \emph{exact}: the model's metres are its own focal times a camera-free output, so the given-camera depth follows exactly from the audit; \emph{rerun}: run again with the camera as input, on the same images as the plain run.}\label{tab:camera}
\setlength{\tabcolsep}{3.5pt}
\begin{tabular}{llrcccc}\toprule
& & & \multicolumn{2}{c}{own focal} & \multicolumn{2}{c}{true camera} \\ \cmidrule(lr){4-5}\cmidrule(lr){6-7}
Model & how & images & $\alpha$ & scale & $\alpha$ & scale \\ \midrule
Depth Pro & exact & 800 & 0.31 & 0.136 & \ensuremath{-0.03} & 0.189 \\
DA3-Nested & exact & 200 & 0.56 & 0.099 & \ensuremath{-0.04} & 0.309 \\
MoGe-2-S & rerun & 800 & 0.36 & 0.214 & \ensuremath{-0.08} & 0.146 \\
UniDepthV2-S & rerun & 800 & 0.34 & 0.138 & 0.13 & 0.104 \\
\bottomrule\end{tabular}\end{table}

\paragraph{Undoing the shrinkage without labels.}
The true camera is rarely available, but the shrinkage of \cref{sec:law} can be inverted without it.
\emph{FocalUnshrink} measures $\hat\beta$ by self-zoom and $\mu$ as the mean unzoomed log focal on unlabeled
images of the target domain, and maps every focal estimate to $\mu+(\log\hat f-\mu)/\hat\beta$. It needs no
training, labels, camera or zoom factor (cross-fitted over image halves; supplement). For Depth Pro
($\hat\beta=\DepthproUnshrinkBeta$, a slope pooled over images; per-image median \DepthproBeta{} in \cref{tab:audit}) it brings $\alpha$ from \DepthproUnshrinkAlpha{} to
\DepthproUnshrinkAlphacorr{} and the scale error at $3\times$ from \DepthproUnshrinkScaleThree{} to
\DepthproUnshrinkScaleThreecorr, while the unzoomed error stays at \DepthproUnshrinkScaleOnecorr{}
(\DepthproUnshrinkScaleOne{} before). DA3-Nested, with a less reliable cue ($\hat\beta=\DaThreeNestedUnshrinkBeta$),
also becomes equivariant ($\alpha=\DaThreeNestedUnshrinkAlphacorr$), but the amplified noise raises its errors,
as the trade-off predicts.

\paragraph{Pixels-only models follow magnification; camera heads differ (Q3).}
\Cref{tab:cues} splits a $2\times$ zoom into its two parts (\cref{sec:cues}). For the pixels-only models,
removing the context around the crop barely changes its depth (DA-V2-S \DavTwoSCueContext, ZoeDepth
\ZoedepthCueContext), while showing the same content $2\times$ larger accounts for almost the whole zoom effect
(\DavTwoSCueMagnify, \ZoedepthCueMagnify). This holds on every dataset: magnification lowers their depth in
\cueMagnifyBelowOne{} of \cuePixPairs{} model--dataset pairs (factors \cueMagnifyMin--\cueMagnifyMax), whereas
the context effect changes sign between datasets (\cueContextMin--\cueContextMax). Shrinking the whole image
raises their depth (DA-V2-S \DavTwoSCueShrink, against 2 for apparent size alone): the apparent-size reading of
\cref{sec:law} in isolation. The camera-predicting models do not share one
mechanism. MoGe-2-S behaves like the pixels-only models (magnify \MogeTwoSCueMagnify, below 1 on
\MogeTwoSCueMagnifyBelowOne{} of 4 datasets). UniDepthV2-S reacts to the lost context instead (context
\UnidepthTwoSCueContext, below 1 on \UnidepthTwoSCueContextBelowOne{} of 4; magnify \UnidepthTwoSCueMagnify).
Depth Pro follows magnification on KITTI and NYUv2 but not on DIODE (pooled \DepthproCueMagnify); DA3-Metric-L
is discussed in the supplement.
\begin{table}[t]\centering\small
\caption{Splitting a $2\times$ zoom into its two parts, median depth ratio on the same pixels (1 = unchanged). \emph{Context}: the crop pasted at its own place and size on a plain canvas. \emph{Magnify}: the zoomed image against that pasted crop. \emph{Zoom}: both together. \emph{Shrink}: the whole image made $2\times$ smaller on a canvas. A model that reads apparent size gives 1, 0.5, 0.5 and 2. $^{*}$Predicted focal on the shrunk image (correct: 0.5). Lower block: DA-V2-S at the fine-tuning input size, frozen and tuned with EFT+B and EFT-S (\cref{sec:eft}).}\label{tab:cues}
\setlength{\tabcolsep}{3.5pt}
\begin{tabular}{lccccc}\toprule
Model & context & magnify & zoom & shrink & focal$^{*}$ \\ \midrule
DA-V2-S & 0.96 & 0.58 & 0.55 & 1.76 & -- \\
ZoeDepth & 0.95 & 0.66 & 0.64 & 1.47 & -- \\
MoGe-2-S & 1.06 & 0.75 & 0.79 & 1.32 & 0.75 \\
UniDepthV2-S & 0.75 & 1.07 & 0.79 & 1.09 & 0.64 \\
Depth Pro & 1.04 & 0.84 & 0.84 & 1.16 & 0.70 \\
DA3-Metric-L & 0.79 & 1.34 & 1.03 & 0.78 & -- \\
\midrule
frozen & 1.01 & 0.54 & 0.56 & 1.79 & -- \\
EFT+B & 1.03 & 0.86 & 0.89 & 1.52 & -- \\
EFT-S & 1.17 & 0.71 & 0.87 & 1.88 & -- \\
\bottomrule\end{tabular}\end{table}

\paragraph{What the label-free score is good for.}
Across the \proxyPairs{} model--dataset pairs, $|\alpha|$ rank-correlates with the error a $3\times$ zoom adds
(Spearman \proxyRhoDeltaThreeTwoWay, [\proxyRhoDeltaThreeTwoWayLo, \proxyRhoDeltaThreeTwoWayHi] resampling models
and datasets), not with unzoomed accuracy (\proxyRhoOne): it tells which model will degrade most under zoom.
When the zoom factor is known, scaling depth by $s^{\hat\alpha}$ (ZoomFix) removes 38--70\% of the error a
$3\times$ zoom adds to DA-V2-S, except on DIODE-outdoor (supplement).

\section{Equivariance fine-tuning}
\label{sec:eft}

\begin{table*}[t]\centering\scriptsize
\caption{Fine-tuning DA-V2-S on unlabeled images, identical preprocessing for all variants. For each: $\alpha$, AbsRel on unzoomed images, and the depth change under the blur control (the resolution shortcut). $^\dagger$never used in training.}\label{tab:eft}
\setlength{\tabcolsep}{2.5pt}
\begin{tabular}{l ccc ccc ccc ccc ccc ccc}\toprule
& \multicolumn{3}{c}{Frozen} & \multicolumn{3}{c}{Zhong et al.-style} & \multicolumn{3}{c}{Blur-only} & \multicolumn{3}{c}{EFT} & \multicolumn{3}{c}{EFT+B (ours)} & \multicolumn{3}{c}{EFT+B, $\lambda{=}3$} \\
\cmidrule(lr){2-4}\cmidrule(lr){5-7}\cmidrule(lr){8-10}\cmidrule(lr){11-13}\cmidrule(lr){14-16}\cmidrule(lr){17-19}
Dataset & $\alpha$ & Abs$_{1\times}$ & blur & $\alpha$ & Abs$_{1\times}$ & blur & $\alpha$ & Abs$_{1\times}$ & blur & $\alpha$ & Abs$_{1\times}$ & blur & $\alpha$ & Abs$_{1\times}$ & blur & $\alpha$ & Abs$_{1\times}$ & blur \\ \midrule
KITTI & 1.03 & 0.138 & 1.1\% & 0.18 & 0.222 & 25.8\% & 0.82 & 0.160 & 0.2\% & 0.18 & 0.156 & 29.6\% & 0.05 & 0.175 & 4.9\% & 0.07 & 0.173 & 6.7\% \\
NYUv2 & 0.98 & 0.344 & 0.7\% & 0.11 & 0.453 & 4.1\% & 0.56 & 0.384 & 0.2\% & 0.12 & 0.313 & 4.2\% & 0.07 & 0.301 & 0.4\% & 0.05 & 0.360 & 0.3\% \\
DIODE-in$^\dagger$ & 0.70 & 0.280 & 1.0\% & 0.39 & 0.419 & 0.6\% & 0.44 & 0.404 & 0.3\% & 0.39 & 0.368 & 0.6\% & 0.37 & 0.400 & 0.4\% & 0.35 & 0.432 & 0.4\% \\
DIODE-out$^\dagger$ & 0.65 & 0.731 & 0.6\% & 0.37 & 0.926 & 1.6\% & 0.45 & 0.802 & 0.2\% & 0.39 & 0.877 & 1.5\% & 0.40 & 1.000 & 0.4\% & 0.41 & 0.971 & 0.4\% \\
\bottomrule\end{tabular}\end{table*}

The law in \cref{eq:law} is not only a test; it is free supervision. Following the
self-consistency idea of Zhong \etal~\cite{zhong2023equivariance}, but with a metric target and a
frozen teacher, for a pixels-only
model $\hat D_\theta$ we keep a frozen copy $\hat D_0$ (the teacher) and minimise, on
unlabeled images $I$,
\begin{equation}
\begin{aligned}
  \mathcal{L} = {} & \big\| \mathcal{U}_b \log \hat D_\theta(\mathcal{Z}_b I) - \log \hat D_0(I)\big|_b \big\|_1 \\
  & + \lambda\, \big\| \log \hat D_\theta(I) - \log \hat D_0(I)\big\|_1 ,
\end{aligned}
  \label{eq:eft}
\end{equation}
with a zoom term (first) and an anchor (second), where $b$ is a random box with zoom factor $s\sim\operatorname{LogUniform}[1,3]$ and random
centre, $\mathcal{U}_b$ resamples the zoomed prediction onto the crop grid, and
$\cdot|_b$ crops. The teacher's prediction on the \emph{unzoomed} image is the target for
the zoomed one, so the student has to discover from the pixels that a zoom happened and
undo the size-based shortcut.
No depth labels and no intrinsics are used.

\paragraph{Setup.}
We tune Depth Anything V2 metric Small (24.8M parameters). The outdoor head is tuned on the
frames of seven KITTI raw drives that do not overlap the 13 drives of the KITTI val selection (3{,}323
images), and the indoor head on the RGB of the 795 official NYUv2 training images. AdamW, learning rate $5\cdot10^{-5}$ (head) and
$5\cdot10^{-6}$ (backbone), batch 4, 3{,}000 steps, $\lambda=1$, horizontal flips, input height
392\,px, mixed precision on one T4 GPU. DIODE is never used for training. The frozen baseline is re-evaluated through the same fixed 392-px input
as the tuned models, so its $\alpha$ and error can differ from \cref{tab:audit}.

\paragraph{The resolution shortcut and EFT+B.}
A digital zoom also removes detail, so a student trained with \cref{eq:eft} alone can learn to detect a
zoom from \emph{sharpness} instead of geometry. The blur control of \cref{sec:audit} exposes exactly this (\cref{sec:eftresults}).
EFT+B adds a third term that removes the shortcut:
\begin{equation}
  \mathcal{L}_{\text{B}} = \mathcal{L} + \gamma\,\big\| \log \hat D_\theta(\mathcal{B}_{s'} I) - \log \hat D_0(I)\big\|_1,
\end{equation}
where $\mathcal{B}_{s'}$ down-samples by an independent $s'\sim\operatorname{LogUniform}[1,3]$ and
up-samples back. A blurred image has the field of view of the original, so its target is the
unzoomed depth; detail loss alone is no longer evidence of a zoom. We use $\gamma=1$.

\paragraph{EFT-S: correcting only the scale.}
The zoom error of the frozen model is almost one number per image: at $3\times$ the log depth of the crop
moves by \fixedShiftThree, its shape (\cref{sec:audit}) by only \fixedShapeThree. EFT-S keeps DA-V2-S frozen and
learns one log-scale per image, $\hat D = \hat D_0(I)\,e^{c(I)}$, with the EFT+B objective; $c$ is a small MLP on
the frozen tokens that trains from a cache on a CPU (supplement).

\paragraph{Baselines and ablations.} Anchor-only tuning is no control (the student starts as the teacher, so
its loss and gradient are exactly zero). \emph{Blur-only} drops the zoom term; a Zhong \etal-style~\cite{zhong2023equivariance}
variant uses the student's own stop-gradient prediction as target.

\section{Results: correcting the models}
\label{sec:eftresults}

\paragraph{Plain self-consistency learns a shortcut.}
EFT lowers $\alpha$ of DA-V2-S from \DavTwoSFixedKittiAlpha{} to \DavTwoSEftKittiAlpha{} on KITTI and
from \DavTwoSFixedNyuAlpha{} to \DavTwoSEftNyuAlpha{} on NYUv2 (\cref{tab:eft}). The blur control reveals how:
the tuned model moves depth by \DavTwoSEftKittiBlur\% on KITTI when the image is only blurred, against
\DavTwoSFixedKittiBlur\% before tuning. It has learned that a blurry image is a zoomed one.

\paragraph{EFT+B removes it.}
With the blur-invariance term the shortcut is gone (\DavTwoSEftbKittiBlur\% on KITTI,
\DavTwoSEftbNyuBlur\% on NYUv2) and $\alpha$ drops further, to \DavTwoSEftbKittiAlpha{} and
\DavTwoSEftbNyuAlpha. On the two datasets never used in training it is \DavTwoSEftbDiodeInAlpha{} and
\DavTwoSEftbDiodeOutAlpha{} (frozen: \DavTwoSFixedDiodeInAlpha{} and \DavTwoSFixedDiodeOutAlpha). Over \eftbSeeds{} training seeds
$\alpha$ is $\EftbSeedsKittiAlphaMean\pm\EftbSeedsKittiAlphaSd$, $\EftbSeedsNyuAlphaMean\pm\EftbSeedsNyuAlphaSd$,
$\EftbSeedsDiodeInAlphaMean\pm\EftbSeedsDiodeInAlphaSd$ and $\EftbSeedsDiodeOutAlphaMean\pm\EftbSeedsDiodeOutAlphaSd$
(mean $\pm$ sd, same order).

\paragraph{The cost.}
AbsRel on unzoomed KITTI images rises from \DavTwoSFixedKittiAbsrelOne{}
to \DavTwoSEftbKittiAbsrelOne{}, and on DIODE-outdoor from \DavTwoSFixedDiodeOutAbsrelOne{} to
\DavTwoSEftbDiodeOutAbsrelOne{} (higher for every seed), while NYUv2 is about unchanged
(\DavTwoSFixedNyuAbsrelOne{} to $\EftbSeedsNyuAbsrelOneMean\pm\EftbSeedsNyuAbsrelOneSd$ over seeds). A stronger
anchor does not buy it back (supplement).

\paragraph{What does the work.}
Blur augmentation alone leaves most of the effect in place ($\alpha$ \DavTwoSBlurKittiAlpha{} on KITTI,
\DavTwoSBlurNyuAlpha{} on NYUv2). The student's own stop-gradient target, as in
Zhong \etal~\cite{zhong2023equivariance}, reaches a similar $\alpha$ (\DavTwoSZhongKittiAlpha,
\DavTwoSZhongNyuAlpha) but drifts: the blur shortcut returns (\DavTwoSZhongKittiBlur\% on KITTI) and unzoomed
AbsRel degrades most (\DavTwoSZhongKittiAbsrelOne{} on KITTI). Teacher, zoom term and blur term are each needed.

\paragraph{Correcting only the scale.}
EFT-S, which learns one number per image, reaches $\alpha=\DavTwoSScaleNyuAlpha$ on NYUv2 at no unzoomed cost
(AbsRel \DavTwoSScaleNyuAbsrelOne, frozen \DavTwoSFixedNyuAbsrelOne). It also transfers better than EFT+B to the unseen
DIODE ($\alpha=\DavTwoSScaleDiodeInAlpha$ and \DavTwoSScaleDiodeOutAlpha, against \DavTwoSEftbDiodeInAlpha{} and
\DavTwoSEftbDiodeOutAlpha), but is weaker on KITTI ($\alpha=\DavTwoSScaleKittiAlpha$) and costs more unzoomed accuracy there and on DIODE-outdoor. Tuning only Depth Pro's focal head (EFT-F)
helps on its training domains, not on DIODE (supplement).

\section{Limitations and conclusion}
\label{sec:conclusion}

\paragraph{Limitations.}
Real optical zoom is tested on only \srrawSequences{} scenes from one lens. When the camera is not given, a crop
also removes surrounding context, so $\alpha=0$ is a goal, not a promise; the test measures how far a model is
from the law and in which direction. Giving the model the true camera is an exact experiment only for Depth Pro
and DA3-Nested. EFT+B lowers accuracy on normal and unseen photos. FocalUnshrink is a first-order correction; it
needs unlabeled photos from the target domain to set its scale, and a reliable focal cue.

\paragraph{Conclusion.}
In short: today's metric depth models that are not told the camera mistake a digital zoom for moving closer,
and their focal-length estimates stay too close to the cameras they were trained on. Our zoom test needs no
labels and no new data, takes only a crop and a resize, and should be a routine check for any model that
reports depth in metres.

\clearpage
\section*{LLM Usage Statement}
Claude (Anthropic) and ChatGPT (OpenAI) were used throughout this work. Working as agents, with limited
step-by-step direction, they wrote and ran the experiment code, analysed the results, produced the figures,
drafted and edited the text, and found and summarised related work and citations. They also proposed research
ideas, such as which models to evaluate and which controls to run, from which the authors chose. No LLM was used
to create or annotate data. The authors checked all code, results, references and claims, and take full
responsibility for the paper.
{
    \small
    \bibliographystyle{ieeenat_fullname}
    \bibliography{main}
}
\clearpage
\appendix
\setcounter{table}{0}\setcounter{figure}{0}
\renewcommand{\thetable}{S\arabic{table}}
\renewcommand{\thefigure}{S\arabic{figure}}
\begin{center}{\Large\bf Supplementary Material}\end{center}
\sloppy

This supplementary material is referred to as ``the supplement'' in the main text. Section and table numbers
without a letter refer to the main paper. Every number here is generated from the same result files as the main paper.

\section{Relation to the closest work}\label{sec:supp_related}

\Cref{tab:related} sets this audit next to the three works closest to it, each described by what it states
about itself. The difference is the setting: the earlier works ask whether a depth \emph{map} follows a crop, or
how a camera-conditioned model degrades when a pipeline passes \emph{stale} intrinsics after a crop or resize.
We give every model the \emph{correct} zoomed camera, or let it infer its own, and ask how its metric scale and
its predicted focal respond, which is what the exponents $\alpha$ and $\beta$, the line $\alpha+\beta=1$ and the
shrinkage account of $\beta$ describe. The stale-$K$ case is one row of our audit
(\cref{sec:law}, $\alpha_{\text{stale}}=\alpha_{\text{true}}+1$).

\begin{table*}[t]\centering\small
\caption{The closest work and this audit, side by side. Each entry summarises the setting the work itself
describes; a dash: not described as part of that work.}\label{tab:related}
\setlength{\tabcolsep}{4pt}
\begin{tabular}{p{2.4cm}p{3.2cm}p{3.0cm}p{3.3cm}p{4.0cm}}\toprule
& Zhong \etal~\cite{zhong2023equivariance} & UniDepth~\cite{piccinelli2024unidepth} & Kim and Lee~\cite{kim2026mismatch} & this work \\ \midrule
Transform & crop-and-resize of the input & rescaled and cropped views in training & crop, resize, padding in preprocessing & crop-and-resize as an exact camera change (\cref{eq:law}) \\
Camera after the crop & does not enter & conditions the features & stale (not updated) & correct $K'$, or inferred by the model \\
What is measured & equivariance of depth and normal maps & (training loss) & bias from the mismatch, focal vs principal-point part & metric scale ($\alpha$) and predicted focal ($\beta$) per image, against the exact zoomed camera \\
Models & supervised depth and normal predictors & UniDepth & camera-conditioned depth & pixels-only, camera-predicting and camera-taking metric models (7 main, 11 more) \\
Explanation & -- & -- & resize scales the focal, crop shifts the principal point & zoom--squash ambiguity, $\alpha+\beta=1$, focal head as shrinkage ($\beta$ = cue reliability) \\
Real optical zoom & -- & -- & -- & SR-RAW, \srrawSequences{} sequences \\
Correction & test-time averaging; self-consistency fine-tuning & invariance loss & Mismatch-Aware Camera Module (uses preprocessing metadata) & FocalUnshrink (no training or labels); EFT+B, EFT-S (unlabeled images) \\
\bottomrule\end{tabular}\end{table*}

\section{Qualitative example on NYUv2}

\begin{figure}[h]
  \centering
  \maybefig{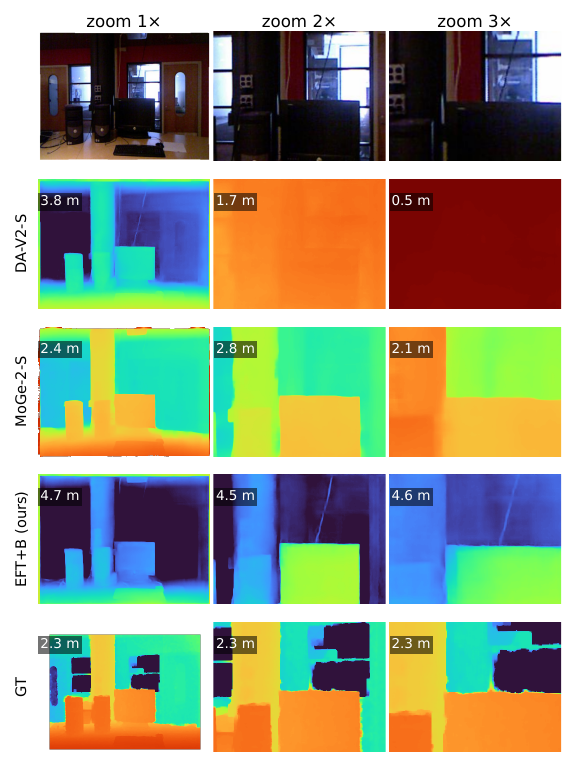}
  \caption{The layout of \cref{fig:teaser} on an NYUv2 test image: depth at $1\times$, $2\times$ and
  $3\times$ digital zoom on one colour scale in metres. The number in each panel is the median depth over
  the pixels inside the $3\times$ crop, the same physical points in every column. EFT+B (392-px input) keeps
  4.65, 4.47 and 4.55\,m (frozen model at that input: 4.16 to 0.81\,m); the ground truth is 2.26\,m.}
  \label{fig:supp_nyu}
\end{figure}

\Cref{fig:supp_nyu} repeats \cref{fig:teaser} on an NYUv2 test image. Depth Anything V2 again moves
the scene closer as the zoom grows (3.85, 1.66 and 0.52\,m); EFT+B keeps the depth of the same points, at a wrong
scale.

\section{The depth exponent per dataset}

\begin{figure}[t]
  \centering
  \maybefig{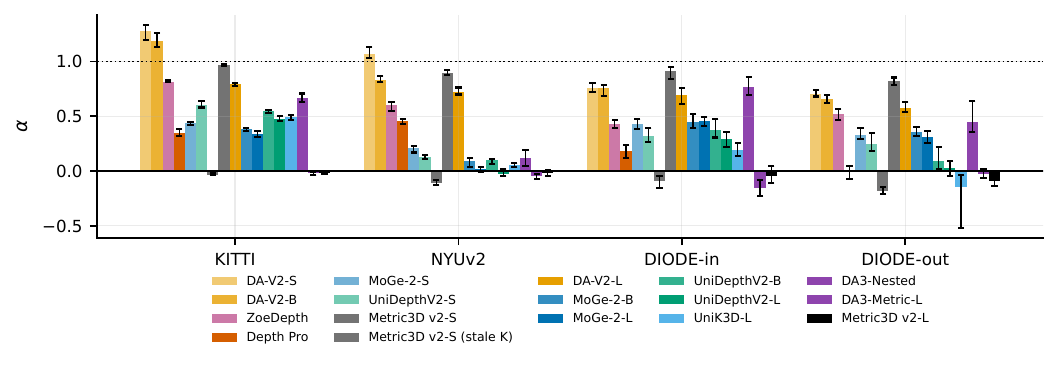}
  \caption{Median $\alpha$ per model and dataset.}
  \label{fig:supp_alpha_ds}
\end{figure}

\Cref{tab:alpha_ds} and \cref{fig:supp_alpha_ds} split the pooled $\alpha$ of \cref{tab:audit} by
dataset. On every set each pixels-only model has a larger $\alpha$ than each camera-predicting model
(ZoeDepth and MoGe-2-S tie on DIODE-indoor), and Metric3D v2 with the true camera stays near zero. The
pixels-only models shrink most on KITTI and NYUv2, where DA-V2-S even exceeds pure apparent-size
behaviour ($\alpha>1$).

\begin{table}[t]\centering\small
\caption{Depth exponent $\alpha$ per dataset (median over images).}\label{tab:alpha_ds}
\setlength{\tabcolsep}{3pt}
\resizebox{\linewidth}{!}{\begin{tabular}{lcccc}\toprule
Model & KITTI & NYUv2 & DIODE-in & DIODE-out \\ \midrule
DA-V2-S & 1.28 & 1.07 & 0.76 & 0.71 \\
DA-V2-B & 1.19 & 0.83 & 0.75 & 0.66 \\
ZoeDepth & 0.82 & 0.60 & 0.43 & 0.52 \\
Depth Pro & 0.35 & 0.46 & 0.19 & \ensuremath{-0.01} \\
MoGe-2-S & 0.44 & 0.21 & 0.43 & 0.33 \\
UniDepthV2-S & 0.60 & 0.13 & 0.32 & 0.25 \\
Metric3D v2-S & \ensuremath{-0.03} & \ensuremath{-0.11} & \ensuremath{-0.09} & \ensuremath{-0.18} \\
Metric3D v2-S (stale K) & 0.97 & 0.89 & 0.91 & 0.82 \\
\bottomrule\end{tabular}}\end{table}

\section{Complete test sets}\label{sec:supp_full}

The main audit uses 200 images per set. \Cref{tab:full} reruns the six main models on every image of the four
sets. Per model and dataset, $\alpha$ moves by at most \fullMaxShift{} (median \fullMedianShift), and the ordering
of pixels-only above camera-predicting models holds on every set (closest: ZoeDepth and MoGe-2-S on DIODE-indoor,
tied on the sample, \ZoedepthFullDiodeInAlpha{} and \MogeTwoSFullDiodeInAlpha{} on the complete set).

\begin{table}[t]\centering\small
\caption{Depth exponent $\alpha$ on the 200-image sample per set / on the complete test sets (KITTI val selection, NYUv2 test, DIODE val indoor and outdoor; \FullKittiImages, \FullNyuImages, \FullDiodeInImages{} and \FullDiodeOutImages{} images), median over images.}\label{tab:full}
\setlength{\tabcolsep}{3pt}
\resizebox{\linewidth}{!}{\begin{tabular}{lcccc}\toprule
Model & KITTI & NYUv2 & DIODE-in & DIODE-out \\ \midrule
DA-V2-S & 1.28 / 1.26 & 1.07 / 1.08 & 0.76 / 0.75 & 0.71 / 0.70 \\
ZoeDepth & 0.82 / 0.82 & 0.60 / 0.59 & 0.43 / 0.43 & 0.52 / 0.50 \\
Depth Pro & 0.35 / 0.33 & 0.46 / 0.44 & 0.19 / 0.20 & \ensuremath{-0.01} / 0.03 \\
MoGe-2-S & 0.44 / 0.43 & 0.21 / 0.18 & 0.43 / 0.42 & 0.33 / 0.34 \\
UniDepthV2-S & 0.60 / 0.60 & 0.13 / 0.11 & 0.32 / 0.30 & 0.25 / 0.21 \\
Metric3D v2-S & \ensuremath{-0.03} / \ensuremath{-0.03} & \ensuremath{-0.11} / \ensuremath{-0.11} & \ensuremath{-0.09} / \ensuremath{-0.08} & \ensuremath{-0.18} / \ensuremath{-0.18} \\
\bottomrule\end{tabular}}\end{table}

\section{How well the power law fits}

Per image, the median RMS residual of $\log$ depth ratio around $-\alpha\log s$ is \fitRmsMedian{} over the main
models. The exponent is not constant over $s$: fitted on $s\le1.5$ and on $s\ge2$ separately, it is
\DavTwoSAlphaSmall{} and \DavTwoSAlphaLarge{} for DA-V2-S, \DepthproAlphaSmall{} and \DepthproAlphaLarge{} for
Depth Pro, \MogeTwoSAlphaSmall{} and \MogeTwoSAlphaLarge{} for MoGe-2-S, and \UnidepthTwoSAlphaSmall{} and
\UnidepthTwoSAlphaLarge{} for UniDepthV2-S (medians over images). The camera-predicting models follow small zooms
better than large ones; a single $\alpha$ over $[1,3]$ averages these.

\Cref{tab:uncert} explains why. With scene-bootstrap intervals for every exponent, the focal exponent of the
camera-predicting models falls from the small to the large zooms (Depth Pro \DepthproUncBetaSmall{} to
\DepthproUncBetaLarge, MoGe-2-S \MogeTwoSUncBetaSmall{} to \MogeTwoSUncBetaLarge), while $\alpha+\beta$ stays near 1
in both ranges. The focal head saturates on large zooms, as a shrinkage estimator does in the tails of its prior,
and the depth follows it along the line. For the pixels-only DA-V2-S, $\alpha$ is close to 1 in both ranges
(\DavTwoSUncAlphaSmall, \DavTwoSUncAlphaLarge), and for Metric3D v2 with the true camera close to 0
(\MetricThreedSUncAlphaSmall, \MetricThreedSUncAlphaLarge).

\begin{table*}[t]\centering\small
\caption{The exponents with 95\% intervals from a scene-level bootstrap (median over images, centred zooms), and fitted separately on the small zooms ($s\le1.5$) and the large ones ($s\ge2$). For the camera-predicting models $\alpha$ grows with the zoom because $\beta$ falls, while $\alpha+\beta$ stays close to 1 in both ranges.}\label{tab:uncert}
\setlength{\tabcolsep}{4pt}
\begin{tabular}{lrcccccc}\toprule
& & \multicolumn{3}{c}{all zooms} & \multicolumn{3}{c}{small / large zooms} \\ \cmidrule(lr){3-5}\cmidrule(lr){6-8}
Model & images & $\alpha$ & $\beta$ & $\alpha+\beta$ & $\alpha$ & $\beta$ & $\alpha+\beta$ \\ \midrule
DA-V2-S & 800 & 0.95 [0.83, 1.12] & -- & -- & 0.85 / 0.97 & -- / -- & -- / -- \\
ZoeDepth & 800 & 0.61 [0.53, 0.75] & -- & -- & 0.68 / 0.60 & -- / -- & -- / -- \\
Depth Pro & 800 & 0.31 [0.20, 0.41] & 0.65 [0.58, 0.75] & 0.97 [0.95, 1.00] & 0.10 / 0.32 & 0.83 / 0.65 & 0.93 / 0.97 \\
MoGe-2-S & 800 & 0.36 [0.30, 0.40] & 0.56 [0.50, 0.63] & 0.95 [0.91, 0.98] & 0.27 / 0.37 & 0.65 / 0.55 & 0.92 / 0.92 \\
UniDepthV2-S & 800 & 0.34 [0.23, 0.51] & 0.45 [0.37, 0.67] & 0.87 [0.79, 0.94] & 0.29 / 0.36 & 0.50 / 0.45 & 0.79 / 0.81 \\
DA3-Nested & 200 & 0.56 [0.44, 0.66] & 0.39 [0.27, 0.56] & 0.96 [0.92, 0.98] & 0.42 / 0.57 & 0.55 / 0.40 & 0.96 / 0.96 \\
Metric3D v2-S & 800 & \ensuremath{-0.07} [\ensuremath{-0.11}, \ensuremath{-0.05}] & -- & -- & \ensuremath{-0.08} / \ensuremath{-0.07} & -- / -- & -- / -- \\
\bottomrule\end{tabular}\end{table*}

\section{A calibration-only reference}\label{sec:supp_geocalib}

GeoCalib~\cite{veicht2024geocalib} predicts the camera but no depth. Run on the same zooms of the same
\geocalibImages{} images, its focal exponent is $\beta=\geocalibBeta$ (median over images; \GeocalibKittiBeta{} on
KITTI, \GeocalibNyuBeta{} on NYUv2, \GeocalibDiodeInBeta{} on DIODE-in, \GeocalibDiodeOutBeta{} on DIODE-out),
against the ideal $\beta=1$, and its median focal error on the unzoomed images is a factor \geocalibFocalErr. A model
built only for calibration also under-scales the focal under zoom, so the under-scaling of the depth models' focal
heads ($\beta=\DepthproBeta$ for Depth Pro, \MogeTwoSBeta{} for MoGe-2-S) is not specific to depth estimation.

\paragraph{A calibrator's focal in place of the true one.} The true camera is rarely available, so a natural
alternative is to give a depth model the focal of a dedicated calibrator. For Depth Pro this is exact without a new
run: its metres are its focal times a camera-free output, so replacing its focal by GeoCalib's rescales each depth map
by $f_{\text{GeoCalib}}/\hat f$. On the \depthproGeocalibImages{} images of the main sample this gives
$\alpha=\depthproGeocalibAlpha$ (own focal \DepthproCamOwnAlpha, true focal \DepthproCamGivenAlpha, FocalUnshrink
\DepthproUnshrinkAlphacorr) and an unzoomed scale error of \depthproGeocalibScale{} (own focal \DepthproCamOwnScale).
GeoCalib's focal under-scales under zoom as well, so it does not repair the zoom response, and it is less accurate
for Depth Pro's depth than Depth Pro's own focal.

\paragraph{Own versus true focal, per dataset.} The pooled unzoomed scale errors of \cref{tab:camera} depend on the
mix of datasets. Per dataset (median $|\log\hat D/D^{*}|$ at $1\times$, own $\to$ true focal), Depth Pro gives
\DepthproCamOwnScaleKitti{} $\to$ \DepthproCamGivenScaleKitti{} on KITTI, \DepthproCamOwnScaleNyu{} $\to$
\DepthproCamGivenScaleNyu{} on NYUv2, \DepthproCamOwnScaleDiodeIn{} $\to$ \DepthproCamGivenScaleDiodeIn{} on
DIODE-indoor and \DepthproCamOwnScaleDiodeOut{} $\to$ \DepthproCamGivenScaleDiodeOut{} on DIODE-outdoor;
DA3-Nested gives \DaThreeNestedCamOwnScaleKitti{} $\to$ \DaThreeNestedCamGivenScaleKitti,
\DaThreeNestedCamOwnScaleNyu{} $\to$ \DaThreeNestedCamGivenScaleNyu, \DaThreeNestedCamOwnScaleDiodeIn{} $\to$
\DaThreeNestedCamGivenScaleDiodeIn{} and \DaThreeNestedCamOwnScaleDiodeOut{} $\to$
\DaThreeNestedCamGivenScaleDiodeOut{} in the same order. The true focal makes both worse on DIODE and leaves NYUv2
about unchanged; it helps Depth Pro on KITTI and hurts DA3-Nested there.

\section{FocalUnshrink}\label{sec:supp_unshrink}

\Cref{tab:unshrink} gives the shrinkage inversion of \cref{sec:law,sec:results} for every focal head. Per model,
$\hat\beta$ is the least-squares slope of $\log\hat f_s/\hat f_1$ on $\log s$ pooled over images (a pooled slope
is more stable than the per-image median of \cref{tab:audit}, \DepthproBeta{} for Depth Pro), and $\mu$ is the mean
unzoomed $\log\hat f$ of the target dataset; both come from one half of each dataset's images and are applied to
the other half, then the halves swap. Without labels the correction cannot remove a model's bias on unzoomed
images: it is centred on the model's own estimate, so at $1\times$ it only spreads the estimates around $\mu$.
What it removes is the error that the zoom adds.

The result follows $\hat\beta$. Depth Pro, with the most reliable focal cue, becomes equivariant and more
accurate under zoom at no cost unzoomed. MoGe-2-S also gains under zoom (its depth is not exactly its focal times
a camera-free output, so its depth columns are an approximation). For UniDepthV2-S and DA3-Nested, with
$\hat\beta\le\UnidepthTwoSUnshrinkBeta$, the $1/\hat\beta$ amplification of per-image noise outweighs the gain,
and for GeoCalib the focal error at $3\times$ falls from \GeocalibUnshrinkFocalThree{} to
\GeocalibUnshrinkFocalThreecorr. Equivariance obtained this way is free of training, but its accuracy is bounded by
how much the camera cue actually knows, which is what $\beta$ measures.

\begin{table*}[t]\centering\small
\caption{FocalUnshrink: the focal head's shrinkage inverted without labels, training or the zoom factor, $\log f_c=\mu+(\log\hat f-\mu)/\hat\beta$ with $\hat\beta$ from self-zoom (pooled over datasets) and $\mu$ the mean unzoomed log focal of the target dataset, both measured on one half of the images and applied to the other (cross-fitted). Before $\to$ after. Scale: median $|\log\hat D/D^{*}|$ on the crop at $1\times$, $3\times$ and over all zooms; focal: median $|\log\hat f/f^{*}|$ at $3\times$. \emph{exact}: metres are the focal times a camera-free output, so the corrected depth is exact; \emph{approx.}: the depth is rescaled as if it were.}\label{tab:unshrink}
\setlength{\tabcolsep}{4pt}
\begin{tabular}{llcccccc}\toprule
Model & depth & $\hat\beta$ & $\alpha$ & scale $1\times$ & scale $3\times$ & scale, all & focal $3\times$ \\ \midrule
Depth Pro & exact & 0.70 & 0.31 $\to$ 0.04 & 0.136 $\to$ 0.138 & 0.467 $\to$ 0.359 & 0.209 $\to$ 0.189 & 0.350 $\to$ 0.360 \\
DA3-Nested & exact & 0.41 & 0.56 $\to$ \ensuremath{-0.05} & 0.099 $\to$ 0.130 & 0.784 $\to$ 0.865 & 0.275 $\to$ 0.368 & 0.559 $\to$ 0.738 \\
MoGe-2-S & approx. & 0.55 & 0.36 $\to$ \ensuremath{-0.08} & 0.214 $\to$ 0.214 & 0.625 $\to$ 0.390 & 0.327 $\to$ 0.281 & 0.573 $\to$ 0.342 \\
UniDepthV2-S & approx. & 0.50 & 0.34 $\to$ \ensuremath{-0.05} & 0.138 $\to$ 0.201 & 0.485 $\to$ 0.458 & 0.211 $\to$ 0.348 & 0.466 $\to$ 0.412 \\
GeoCalib & focal only & 0.58 & -- & -- & -- & -- & 0.453 $\to$ 0.306 \\
\bottomrule\end{tabular}\end{table*}

\paragraph{Sensitivity.} \Cref{tab:unshrink_sens} varies the two measured quantities. $\hat\beta$ is known to
within its scene-bootstrap interval (Depth Pro \DepthproSensBetaCi{} [\DepthproSensBetaLo, \DepthproSensBetaHi]);
at the ends of that interval Depth Pro's corrected $\alpha$ is \DepthproSensAlphaAtHi{} and
\DepthproSensAlphaAtLo, an over- or under-correction of the size the interval implies. The prior centre $\mu$
cancels in the ratio of two zooms, so it cannot affect $\alpha$; taking it from the other datasets instead of the
target one changes only the unzoomed scale (Depth Pro \DepthproSensMuShiftScaleOne, against
\DepthproUnshrinkScaleOnecorr{} with the target's $\mu$). Unlabeled target images matter for scale, not for
equivariance.

\paragraph{Non-Gaussian priors.} The inversion is exact when $\log f$ and the cue are Gaussian. Real camera
distributions are not (phones cluster at a few focal lengths). \Cref{tab:shrink_sim} simulates an exact
posterior-mean focal head under Laplace, uniform and bimodal priors of the same variance. The measured $\hat\beta$
stays close to the Gaussian reliability $\rho$, but the head is no longer linear in the cue, so a single
$\hat\beta$ cannot undo it image by image: the per-image zoom error falls from \SimBimodalRhoSixFiveZerr{} to
\SimBimodalRhoSixFiveZerrc{} for the bimodal prior at $\rho=0.65$, against 0 for the Gaussian one. In every
case the error at $1\times$ grows, by the $1/\hat\beta$ amplification of \cref{sec:law}. FocalUnshrink is therefore
a first-order correction; a non-linear inverse would need the prior's shape, which self-zoom alone does not give.

\begin{table*}[t]\centering\small
\caption{Sensitivity of FocalUnshrink on real data. $\hat\beta$ with a 95\% scene-bootstrap interval; $\alpha$ after the correction with $\hat\beta$ at the upper end of its interval, at its estimate and at the lower end. Scale: median $|\log\hat D/D^{*}|$ at $1\times$, before, after, and after with a misspecified prior centre $\mu$ taken from the \emph{other} datasets instead of the target one. $\mu$ does not enter $\alpha$ (it cancels in the ratio of two zooms), so a wrong prior centre costs scale, not equivariance.}\label{tab:unshrink_sens}
\setlength{\tabcolsep}{4pt}
\begin{tabular}{lccccccc}\toprule
& & \multicolumn{2}{c}{$\alpha$} & \multicolumn{3}{c}{scale $1\times$} & \\ \cmidrule(lr){3-4}\cmidrule(lr){5-7}
Model & $\hat\beta$ [95\%] & before & after ($\hat\beta_{\text{hi}}$ / $\hat\beta$ / $\hat\beta_{\text{lo}}$) & before & after & wrong $\mu$ & $\alpha$, wrong $\mu$ \\ \midrule
Depth Pro & 0.70 [0.62, 0.78] & 0.31 & 0.13 / 0.04 / \ensuremath{-0.09} & 0.136 & 0.138 & 0.245 & 0.03 \\
DA3-Nested & 0.41 [0.29, 0.55] & 0.56 & 0.23 / \ensuremath{-0.05} / \ensuremath{-0.50} & 0.099 & 0.130 & 0.593 & \ensuremath{-0.05} \\
MoGe-2-S & 0.55 [0.51, 0.62] & 0.36 & 0.02 / \ensuremath{-0.08} / \ensuremath{-0.18} & 0.214 & 0.214 & 0.432 & \ensuremath{-0.09} \\
UniDepthV2-S & 0.50 [0.39, 0.64] & 0.34 & 0.14 / \ensuremath{-0.05} / \ensuremath{-0.30} & 0.138 & 0.201 & 0.431 & \ensuremath{-0.07} \\
\bottomrule\end{tabular}\end{table*}

\begin{table}[t]\centering\small
\caption{FocalUnshrink when the camera prior is not Gaussian (simulation, \cref{sec:supp_unshrink}): $\log f$ from a unit-variance prior, a noisy cue of Gaussian reliability $\rho$, the exact posterior-mean focal head. Zoom error: per-image RMS of $\log\hat f_3-\log\hat f_1-\log 3$; error: RMS of $\log\hat f$ against the truth. Before $\to$ after the correction.}\label{tab:shrink_sim}
\setlength{\tabcolsep}{3pt}
\resizebox{\linewidth}{!}{\begin{tabular}{lccccc}\toprule
prior & $\rho$ & $\beta$ & zoom error & error $1\times$ & error $3\times$ \\ \midrule
gaussian & 0.65 & 0.65 & 0.38 $\to$ 0.00 & 0.59 $\to$ 0.74 & 0.70 $\to$ 0.74 \\
gaussian & 0.40 & 0.40 & 0.66 $\to$ 0.00 & 0.78 $\to$ 1.23 & 1.01 $\to$ 1.23 \\
laplace & 0.65 & 0.64 & 0.43 $\to$ 0.28 & 0.57 $\to$ 0.74 & 0.71 $\to$ 0.85 \\
laplace & 0.40 & 0.40 & 0.67 $\to$ 0.37 & 0.76 $\to$ 1.25 & 1.01 $\to$ 1.40 \\
uniform & 0.65 & 0.59 & 0.51 $\to$ 0.41 & 0.58 $\to$ 0.81 & 0.76 $\to$ 0.74 \\
uniform & 0.40 & 0.39 & 0.69 $\to$ 0.30 & 0.77 $\to$ 1.28 & 1.02 $\to$ 1.20 \\
bimodal & 0.65 & 0.58 & 0.54 $\to$ 0.48 & 0.56 $\to$ 0.81 & 0.79 $\to$ 0.77 \\
bimodal & 0.40 & 0.38 & 0.69 $\to$ 0.35 & 0.77 $\to$ 1.28 & 1.04 $\to$ 1.21 \\
\bottomrule\end{tabular}}\end{table}

\section{Objects of familiar size and surfaces}\label{sec:supp_regions}

If a model fixed scale from the usual size of known objects, its depth on those objects should follow the zoom
less than its depth on walls, floors and roads. We split each image with SegFormer-B5 trained on
ADE20K~\cite{xie2021segformer} into object classes (people, cars, chairs, beds, \etc) and surface classes (wall,
floor, road, ceiling, \etc), and fit $\alpha$ per region from the $2\times$ and $3\times$ zooms on the 200
images per set. On the \DavTwoSRegionPairs{} images with both kinds of region, the median $\alpha$ on objects and
surfaces is \DavTwoSRegionObject{} and \DavTwoSRegionSurface{} for DA-V2-S, \DepthproRegionObject{} and
\DepthproRegionSurface{} for Depth Pro, and \MogeTwoSRegionObject{} and \MogeTwoSRegionSurface{} for MoGe-2-S.
Objects follow the zoom slightly \emph{more}, not less (median paired difference \DavTwoSRegionDiff,
\DepthproRegionDiff{} and \MogeTwoSRegionDiff; larger on objects in \DavTwoSRegionObjectLarger\%,
\DepthproRegionObjectLarger\% and \MogeTwoSRegionObjectLarger\% of images; Wilcoxon $p=\DavTwoSRegionP$,
\DepthproRegionP{} and \MogeTwoSRegionP). We find no sign that familiar objects anchor the scale; the difference
is small and the response is nearly uniform over the image.

\section{Bigger and newer models}

\Cref{tab:ladder} extends the audit to larger variants and to Depth Anything~3 (DA3),
UniK3D and VGGT
(100 images per set; 50 for the DA3 and VGGT runs). On the \nLadderShared{} images that every model saw,
size helps the pixels-only family but does not remove the effect: $\alpha$ falls from \DavTwoSAlphaShared{}
(DA-V2-S) to \DavTwoBAlphaShared{} (B) and \DavTwoLAlphaShared{} (L), so the large model still returns less
than half the depth at $3\times$. MoGe-2 is flat across sizes (\MogeTwoSAlphaShared, \MogeTwoBAlphaShared,
\MogeTwoLAlphaShared), with a focal exponent stuck near \MogeTwoLBetaShared. UniDepthV2 improves with size
(\UnidepthTwoSAlphaShared{} to \UnidepthTwoLAlphaShared, $\beta$ from \UnidepthTwoSBetaShared{} to
\UnidepthTwoLBetaShared), and UniK3D-L comes closest
($\alpha=\UnikThreedLAlphaShared$, $\beta=\UnikThreedLBetaShared$). None reaches the level of a model that
is given the correct camera (Metric3D v2, $\alpha=\MetricThreedSAlphaShared$). Scale and newer camera heads
narrow the gap; they do not close it.

Size alone is not the answer. On the \nSmallShared{} images it shares with the main models, the largest
model in the study, DA3-Nested (\DaThreeNestedParams M parameters), shrinks depth with
$\alpha=\DaThreeNestedAlphaSmallShared$ (\DaThreeNestedAlpha{} on all \DaThreeNestedCamImages{} of its images), more than Depth Pro, MoGe-2-S and UniDepthV2-S
(\DepthproAlphaSmallShared, \MogeTwoSAlphaSmallShared, \UnidepthTwoSAlphaSmallShared) and close to the
pixels-only ZoeDepth (\ZoedepthAlphaSmallShared); its focal follows the zoom only partly
($\beta=\DaThreeNestedBetaSmallShared$) and, on its own sample, $\alpha+\beta=\DaThreeNestedAlphaPlusBeta$, again the
apparent-size line. Given the true camera, both DA3-Metric-L and Metric3D v2-L keep depth
($\alpha=\DaThreeMetricLAlphaSmallShared$ and \MetricThreedLAlphaSmallShared). Of the two relative-depth
models, VGGT-1B tracks the focal change well ($\beta=\VggtBetaSmallShared$) and DA3-L does not
(\DaThreeLBetaSmallShared). What decides the behaviour is whether the camera enters the model, not how
large the model is.

\begin{table*}[t]\centering\small
\caption{Model ladder: larger variants and newer families. Left: each model on its own sample (200 images per set for the main models, 100 for the larger variants and UniK3D, 50 for DA3, VGGT and Metric3D v2-L). Right: the models with at least 100 images per set on the \nLadderShared{} images that all of them saw (paired). DA3-L and VGGT normalise depth per scene, so only $\beta$ is defined for them. pred.\ K: predicts the camera; takes K: given the true camera.}\label{tab:ladder}
\setlength{\tabcolsep}{3pt}
\begin{tabular}{llrcccc}\toprule
& & & \multicolumn{2}{c}{own sample} & \multicolumn{2}{c}{shared} \\ \cmidrule(lr){4-5}\cmidrule(lr){6-7}
Model & Group & Params & $\alpha$ & $\beta$ & $\alpha$ & $\beta$ \\ \midrule
DA-V2-S & pixels & 25M & 0.95 & -- & 0.87 & -- \\
DA-V2-B & pixels & 97M & 0.85 & -- & 0.81 & -- \\
ZoeDepth & pixels & 345M & 0.61 & -- & 0.60 & -- \\
Depth Pro & pred. K & 952M & 0.31 & 0.65 & 0.24 & 0.74 \\
MoGe-2-S & pred. K & 35M & 0.36 & 0.56 & 0.37 & 0.53 \\
UniDepthV2-S & pred. K & 34M & 0.34 & 0.45 & 0.35 & 0.39 \\
Metric3D v2-S & takes K & 37M & \ensuremath{-0.07} & -- & \ensuremath{-0.07} & -- \\
DA-V2-L & pixels & 335M & 0.73 & -- & 0.72 & -- \\
MoGe-2-B & pred. K & 105M & 0.35 & 0.62 & 0.37 & 0.55 \\
MoGe-2-L & pred. K & 331M & 0.31 & 0.61 & 0.35 & 0.55 \\
UniDepthV2-B & pred. K & 114M & 0.30 & 0.40 & 0.31 & 0.44 \\
UniDepthV2-L & pred. K & 354M & 0.19 & 0.69 & 0.23 & 0.68 \\
UniK3D-L & pred. K & 359M & 0.16 & 0.77 & 0.17 & 0.82 \\
DA3-Nested & pred. K & 1,690M & 0.56 & 0.39 & -- & -- \\
DA3-Metric-L & takes K & 334M & \ensuremath{-0.05} & -- & -- & -- \\
Metric3D v2-L & takes K & 412M & \ensuremath{-0.03} & -- & -- & -- \\
DA3-L & pred. K (rel.) & 411M & -- & 0.37 & -- & -- \\
VGGT-1B & pred. K (rel.) & 1,257M & -- & 0.81 & -- & -- \\
\bottomrule\end{tabular}\end{table*}

\paragraph{Cue split, further notes.} On the shrunk image, MoGe-2-S, UniDepthV2-S and Depth Pro all shrink their
focal less than the correct factor 0.5 (\MogeTwoSCueShrinkFocal, \UnidepthTwoSCueShrinkFocal,
\DepthproCueShrinkFocal). DA3-Metric-L, which is given the camera, moves in opposite directions under the two
edits of \cref{tab:cues} (\DaThreeMetricLCueContext{} and \DaThreeMetricLCueMagnify), which nearly cancel in the
full zoom (\DaThreeMetricLCueZoom).

\section{What the label-free score is good for}\label{sec:supp_label}

\paragraph{When does the label-free score predict error?}
The \proxyPairs{} model--dataset pairs share models and datasets, so we resample both together for
the intervals. $|\alpha|$ rank-correlates with AbsRel at $3\times$ (Spearman \proxyRhoThree; 95\% interval
[\proxyRhoThreeLo, \proxyRhoThreeHi] when pairs are resampled, but [\proxyRhoThreeTwoWayLo,
\proxyRhoThreeTwoWayHi] when models and datasets are, so this is not established). Part of that association is
mechanical, since a shrinking depth is also a growing error. Subtracting the unzoomed prediction's error on the
same pixels removes the error a model has anyway: $|\alpha|$ then rank-correlates with the added error
(\proxyRhoDeltaThreeTwoWay, [\proxyRhoDeltaThreeTwoWayLo, \proxyRhoDeltaThreeTwoWayHi]). Within KITTI, NYUv2 and
DIODE-indoor the plain correlation is \ProxyKitti, \ProxyNyu{} and \ProxyDiodeIn, and within DIODE-outdoor
\ProxyDiodeOut, where errors are dominated by UniDepthV2. The score does not predict accuracy on unzoomed images
(\proxyRhoOne). We read this narrowly: without labels, the audit tells which model will
degrade most under zoom, not which model is more accurate.

\paragraph{Flagging images to distrust.}
The same score also works per image. For each model and dataset we rank the images by their own
inconsistency at $2\times$ and keep only the most consistent ones. If the score carries information, the
mean AbsRel at $2\times$ of the kept images should fall faster than with a random order. It does in
\riskBelowOne{} of \riskPairs{} model--dataset pairs. On average the area under the risk--coverage curve
is \riskAurc{} of the random baseline (1 = no better than random): \RiskNyu{} on NYUv2, \RiskDiodeIn{} on
DIODE-indoor, and a weaker \RiskKitti{} on KITTI and \RiskDiodeOut{} on DIODE-outdoor. The two pairs above
1 are both on DIODE-outdoor (DA-V2-S and UniDepthV2-S). So a user can run the zoom check on their own photos and set aside the images where the model is
least sure of the scale, without any labels.

\paragraph{A training-free fix when the zoom is known (ZoomFix).}
Phones record a digital zoom ratio, and apps know their own crops. Since $\alpha$ needs no labels, a
user can measure it on their own photos and multiply the depth of zoomed images by $s^{\hat\alpha}$.
We evaluate this cross-fitted: $\hat\alpha$ is measured without labels on half of a dataset's images and
the correction scored on the other half. For DA-V2-S at $3\times$ it removes 61\%, 38\% and 70\% of the error
the zoom adds on KITTI (AbsRel \DavTwoSFixKittiZoomed{} to \DavTwoSFixKittiAbsrel, against
\DavTwoSFixKittiTarget{} for the unzoomed prediction on the same pixels), NYUv2 (\DavTwoSFixNyuZoomed{} to
\DavTwoSFixNyuAbsrel) and DIODE-indoor (\DavTwoSFixDiodeInZoomed{} to \DavTwoSFixDiodeInAbsrel). On
DIODE-outdoor the mean error rises (\DavTwoSFixDiodeOutZoomed{} to \DavTwoSFixDiodeOutAbsrel). The cause is
a heavy tail: even unzoomed, the mean AbsRel there (\DavTwoSFixDiodeOutTarget) is far above the median
(\DavTwoSFixDiodeOutTargetMedian), and scaling depth up amplifies the images that are already
over-estimated, while the typical image improves. Depth Pro, whose $\alpha$ is already small, barely changes. A single global factor cannot undo a zoom that also distorts shape (\cref{sec:audit});
it is a stop-gap for deployed pixels-only models, not a replacement for a camera-aware one.

\section{Depth range of the scene}

A softer reading of the zoom--squash ambiguity would be that scenes with a larger depth range, where the squash
should be more visible, are easier. The within-dataset rank correlation between $\alpha$ and the
inter-quartile range of log depth is \DepthproSpreadRho{} for Depth Pro and \MogeTwoSSpreadRho{} for MoGe-2, so
this is not supported.

\section{Off-centre crops and the principal point}

\begin{figure}[t]
  \centering
  \maybefig{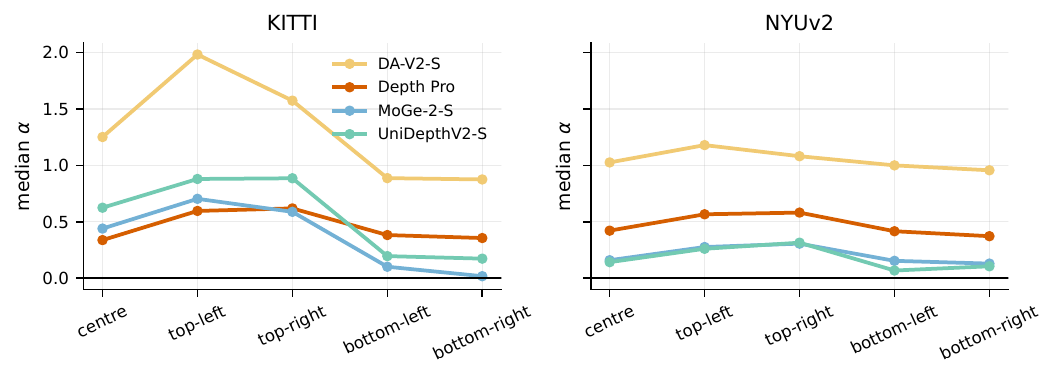}
  \caption{Median $\alpha$ by crop position.}
  \label{fig:supp_offc}
\end{figure}

\begin{figure}[t]
  \centering
  \maybefig{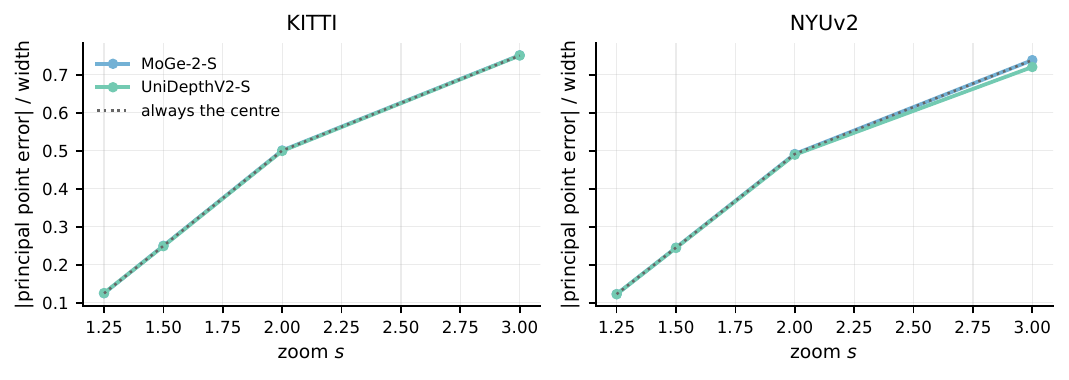}
  \caption{Principal-point error (fraction of the image width) under quadrant zooms. Dotted: the error
  of always answering the image centre.}
  \label{fig:supp_pp}
\end{figure}

Zooms about the four quadrant centres (50 images per set from KITTI and NYUv2) give $\alpha$ of the same
order as centred zooms (\cref{fig:supp_offc}). For such crops the zoom law moves the principal point
by $(c_x-x_0+\tfrac12)\,W/w-\tfrac12$, half the image width at $2\times$. \Cref{fig:supp_pp} compares the
principal point predicted by MoGe-2 and UniDepthV2 with the law: both errors lie on the curve of a
predictor that always answers the image centre (\MogeTwoSPpQuadTwo{} and \UnidepthTwoSPpQuadTwo{} of
the width at $2\times$, against \MogeTwoSPpGuessTwo).

\section{ZoomFix per dataset}

\Cref{tab:zoomfix} gives the cross-fitted ZoomFix of \cref{sec:supp_label} per dataset. The label-free
exponent tracks the dataset: it is above 1 for DA-V2-S on KITTI and NYUv2, where the pooled $\alpha$ hides
a stronger shrink. The MoGe-2-S entry for DIODE-outdoor is dominated by a few zoomed images with very
large errors (the zoomed AbsRel is \MogeTwoSFixDiodeOutZoomed{} against \MogeTwoSFixDiodeOutTarget{}
unzoomed), which a global factor cannot repair.

\begin{table*}[t]\centering\scriptsize
\caption{ZoomFix per dataset at $3\times$, cross-fitted: the exponent $\hat e$ is measured without labels on half of the images and the correction scored on the other half. AbsRel of the zoomed prediction as is, after multiplying by $s^{\hat e}$, and of the unzoomed prediction on the same pixels (the target).}\label{tab:zoomfix}
\setlength{\tabcolsep}{3pt}
\begin{tabular}{l cccc cccc cccc cccc}\toprule
& \multicolumn{4}{c}{KITTI} & \multicolumn{4}{c}{NYUv2} & \multicolumn{4}{c}{DIODE-in} & \multicolumn{4}{c}{DIODE-out} \\
\cmidrule(lr){2-5}\cmidrule(lr){6-9}\cmidrule(lr){10-13}\cmidrule(lr){14-17}
Model & $\hat e$ & as is & fixed & target & $\hat e$ & as is & fixed & target & $\hat e$ & as is & fixed & target & $\hat e$ & as is & fixed & target \\ \midrule
DA-V2-S & 1.25 & 0.758 & 0.384 & 0.149 & 1.12 & 0.675 & 0.495 & 0.205 & 0.75 & 0.598 & 0.365 & 0.263 & 0.69 & 0.558 & 0.661 & 0.605 \\
Depth Pro & 0.31 & 0.258 & 0.277 & 0.160 & 0.44 & 0.449 & 0.223 & 0.121 & 0.19 & 0.399 & 0.421 & 0.201 & 0.12 & 0.636 & 0.638 & 0.665 \\
MoGe-2-S & 0.44 & 0.536 & 0.262 & 0.255 & 0.12 & 0.272 & 0.227 & 0.129 & 0.44 & 0.493 & 0.364 & 0.257 & 0.38 & 3.037 & 4.396 & 0.379 \\
\bottomrule\end{tabular}\end{table*}

\section{EFT-S per domain}

\paragraph{Head.} $c$ is a two-layer MLP (198{,}657 parameters) on the last-layer class token and the mean patch
token of the frozen ViT. Its last layer starts at zero, so training starts from the frozen model, and the depth
shape is the frozen model's by construction.

\paragraph{Training from a cache.} Because $c$ is one number per view, each term of the loss depends only on the
distribution of the residual $\log\hat D_0(\text{view})-\log\hat D_0(I)$ over the crop. We store 100 quantiles of
that residual and the token features for 4{,}000 random zoom and blur views per domain, drawn as for EFT+B
from the same training images, from a single pass of the frozen model. The head is then fitted on this cache
(60 epochs, AdamW, learning rate $10^{-3}$, batch 64) on a CPU.

\paragraph{Results.} \Cref{tab:scale} gives the per-dataset numbers of the scale-only variant EFT-S discussed in
\cref{sec:eftresults}, next to the frozen model and EFT+B. Without the blur term, blur alone moves its KITTI depth by
\DavTwoSScalenbKittiBlur\% (\DavTwoSScaleKittiBlur\% with it): EFT-S finds the same blur shortcut as EFT.

\begin{table*}[t]\centering\scriptsize
\caption{Scale-only equivariance tuning (EFT-S): DA-V2-S stays frozen and a small head learns one log-scale correction per image from its tokens, with the EFT+B objective. $\alpha$, AbsRel on unzoomed images, and the depth change under the blur control. $^\dagger$never used in training.}\label{tab:scale}
\setlength{\tabcolsep}{3pt}
\begin{tabular}{l ccc ccc ccc ccc}\toprule
& \multicolumn{3}{c}{Frozen} & \multicolumn{3}{c}{EFT+B} & \multicolumn{3}{c}{EFT-S (ours)} & \multicolumn{3}{c}{EFT-S, no blur term} \\
\cmidrule(lr){2-4}\cmidrule(lr){5-7}\cmidrule(lr){8-10}\cmidrule(lr){11-13}
Dataset & $\alpha$ & Abs$_{1\times}$ & blur & $\alpha$ & Abs$_{1\times}$ & blur & $\alpha$ & Abs$_{1\times}$ & blur & $\alpha$ & Abs$_{1\times}$ & blur \\ \midrule
KITTI & 1.03 & 0.138 & 1.1\% & 0.05 & 0.175 & 4.9\% & 0.19 & 0.197 & 7.3\% & 0.17 & 0.176 & 23.6\% \\
NYUv2 & 0.98 & 0.344 & 0.7\% & 0.07 & 0.301 & 0.4\% & 0.04 & 0.346 & 0.7\% & 0.11 & 0.347 & 0.8\% \\
DIODE-in$^\dagger$ & 0.70 & 0.280 & 1.0\% & 0.37 & 0.400 & 0.4\% & 0.23 & 0.333 & 1.3\% & 0.26 & 0.335 & 1.1\% \\
DIODE-out$^\dagger$ & 0.65 & 0.731 & 0.6\% & 0.40 & 1.000 & 0.4\% & 0.27 & 1.179 & 1.4\% & 0.25 & 0.940 & 1.4\% \\
\bottomrule\end{tabular}\end{table*}

\section{EFT-F: tuning only the focal head}\label{sec:supp_eftf}

Depth Pro predicts depth as $\hat f\,C(I)/W$, and with the true focal its camera-free part $C$ already follows
a zoom (\cref{tab:camera}), so its zoom error sits in the focal head. A digital zoom by $s$ multiplies the
true focal in pixels by $s$, which supervises that head without labels. EFT-F tunes only the head's layers
after its frozen ViT (the neck, the fusion convolution and the output head) with
\begin{equation*}
\begin{aligned}
L={}&\bigl|\log t(Z_sI)-\log t_0(I)+\log s\bigr|\\
&+\bigl|\log t(BI)-\log t_0(I)\bigr|+\bigl|\log t(I)-\log t_0(I)\bigr|,
\end{aligned}
\end{equation*}
where $t=\tan(\mathrm{fov}/2)$, $t_0$ is the original head, $Z_s$ a random zoom ($1\le s\le3$, random box) and
$B$ a blur with the same field of view, as for EFT+B. The depth network is untouched. Everything the tuned
layers read is frozen, so it is computed once for 4 views of each of 300 unlabeled KITTI raw and 300 NYUv2
train images (10\% held out), and the layers are fitted on that cache (AdamW, learning rate $10^{-4}$, batch
32, \eftfEpochs{} epochs). On held-out zoomed views the focal error drops from \eftfValZoomBefore{} to
\eftfValZoomAfter{} in $\log t$.

\Cref{tab:eftf} gives the audit on the same 200 images per set. On the two training domains EFT-F lowers
$\alpha$ (KITTI \DepthproFheadKittiAlpha{} to \DepthproEftfFheadKittiAlpha, NYUv2 \DepthproFheadNyuAlpha{} to
\DepthproEftfFheadNyuAlpha) and the error at $3\times$ (NYUv2 AbsRel \DepthproFheadNyuAbsrelThree{} to
\DepthproEftfFheadNyuAbsrelThree), at a small cost at $1\times$. It does \emph{not} transfer: on DIODE, which it
never saw, $\alpha$ rises from \DepthproFheadDiodeInAlpha{} to \DepthproEftfFheadDiodeInAlpha{} (indoor) and from
\DepthproFheadDiodeOutAlpha{} to \DepthproEftfFheadDiodeOutAlpha{} (outdoor). A focal head tuned on two domains learns those
domains; the label-free loss allows any unlabeled images, and broader training data is the obvious next step
that we have not run.

\begin{table}[t]\centering\small
\caption{EFT-F: Depth Pro with only its focal head tuned on unlabeled KITTI and NYUv2 training images (the depth network is unchanged). Same 200 images per set; median $\alpha$, $\beta$, mean AbsRel at $1\times$ and $3\times$; ideal $\alpha=0$, $\beta=1$.}\label{tab:eftf}
\footnotesize\setlength{\tabcolsep}{2pt}
\begin{tabular}{lcccccccc}\toprule
& \multicolumn{2}{c}{$\alpha$} & \multicolumn{2}{c}{$\beta$} & \multicolumn{2}{c}{Abs$_{1\times}$} & \multicolumn{2}{c}{Abs$_{3\times}$} \\ \cmidrule(lr){2-3}\cmidrule(lr){4-5}\cmidrule(lr){6-7}\cmidrule(lr){8-9}
Set & DP & EFT-F & DP & EFT-F & DP & EFT-F & DP & EFT-F \\ \midrule
KITTI & 0.35 & 0.23 & 0.63 & 0.75 & 0.161 & 0.184 & 0.251 & 0.192 \\
NYUv2 & 0.46 & 0.09 & 0.56 & 0.93 & 0.101 & 0.113 & 0.464 & 0.247 \\
DIODE-in & 0.19 & 0.44 & 0.73 & 0.44 & 0.194 & 0.233 & 0.386 & 0.411 \\
DIODE-out & \ensuremath{-0.01} & 0.44 & 0.96 & 0.53 & 0.786 & 0.788 & 0.739 & 0.811 \\
\bottomrule\end{tabular}\end{table}

\section{Fine-tuning details}

\paragraph{Training drives.} The training kernel asserts that its seven KITTI drives are disjoint from the 13 drives
of the evaluation selection.

\paragraph{Training seeds.} EFT+B was trained with \eftbSeeds{} seeds (image order, flips, zoom boxes, blur strengths and the torch random
state). Unzoomed AbsRel over seeds is $\EftbSeedsKittiAbsrelOneMean\pm\EftbSeedsKittiAbsrelOneSd$ on
KITTI, $\EftbSeedsNyuAbsrelOneMean\pm\EftbSeedsNyuAbsrelOneSd$ on NYUv2,
$\EftbSeedsDiodeInAbsrelOneMean\pm\EftbSeedsDiodeInAbsrelOneSd$ on DIODE-in and
$\EftbSeedsDiodeOutAbsrelOneMean\pm\EftbSeedsDiodeOutAbsrelOneSd$ on DIODE-out (frozen: \DavTwoSFixedKittiAbsrelOne,
\DavTwoSFixedNyuAbsrelOne, \DavTwoSFixedDiodeInAbsrelOne, \DavTwoSFixedDiodeOutAbsrelOne). The tables report seed 0.

\paragraph{Anchor weight.} A three times stronger anchor gives KITTI AbsRel \DavTwoSEftbThreeKittiAbsrelOne{} and NYUv2
AbsRel \DavTwoSEftbThreeNyuAbsrelOne{} at $\alpha=\DavTwoSEftbThreeKittiAlpha$ and \DavTwoSEftbThreeNyuAlpha: it does not
buy the KITTI accuracy back and costs NYUv2, so the trade-off is not a matter of tuning the weight.

\paragraph{Variants.} All variants start from the same DA-V2-S checkpoint and differ only in the loss:
EFT (anchor + zoom term), EFT+B (+ blur term), EFT+B with $\lambda=3$, blur-only (anchor + blur term),
and the Zhong \etal-style target (anchor + zoom term, with the student's own stop-gradient prediction on
the unzoomed image as target). Each is trained once per domain (KITTI raw outdoor, NYUv2 train indoor)
with the settings of \cref{sec:eft}.

\begin{figure}[t]
  \centering
  \maybefig{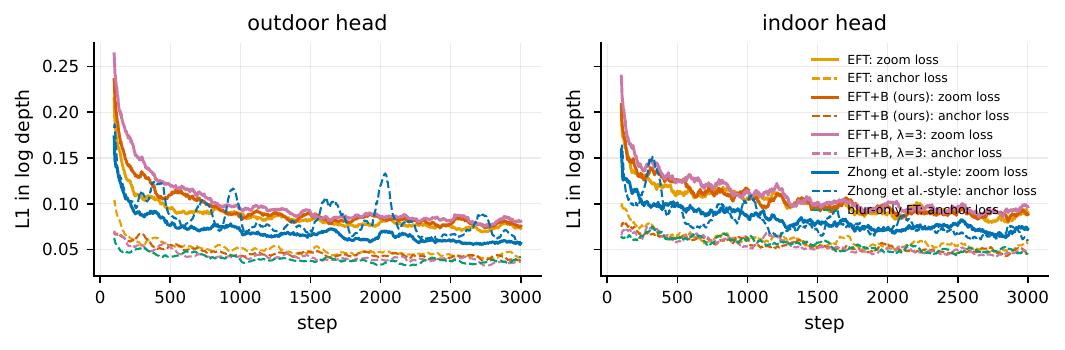}
  \caption{Zoom loss (solid) and anchor loss (dashed) of the fine-tuning variants, outdoor and indoor
  heads. Blur-only has no zoom term.}
  \label{fig:supp_train}
\end{figure}

\paragraph{Training curves.} \Cref{fig:supp_train} shows the zoom and anchor terms over training
(running mean over 100 steps). The anchor term is exactly zero at the first step, since the student is a
copy of the teacher; it jumps as soon as the zoom term moves the weights and then falls through training
(EFT outdoor: mean 0.10 over the first 100 steps, 0.04 over the last 100), as the model learns to
satisfy both terms.

\section{Confounder controls}

\Cref{tab:controls} gives the optical-versus-digital, bilinear and JPEG controls discussed in
\cref{sec:results}. Depth Pro shrinks less with real detail (\DepthproOptical{} optical, \DepthproDigitalLr{}
digital), MoGe-2-S slightly more (\MogeTwoSOptical, \MogeTwoSDigitalLr), and Metric3D v2 with the true $K'$ stays
near 1 under every control (\MetricThreedSOptical, \MetricThreedSDigitalLr, \MetricThreedSJpeg). On
DIODE-outdoor, UniDepthV2-S over-estimates the scale by more than $2\times$ on \udDiodeOutOverTwo\% of the images.

\begin{table}[t]\centering\small
\caption{Confounder controls: median depth ratio on the same pixels (1 = unchanged). Optical: real-detail $2\times$ zoom of a half-resolution camera; digital: the same zoom by crop and up-sampling; bilinear: the $2\times$ digital zoom with bilinear resampling; JPEG: quality 75, no zoom.}\label{tab:controls}
\setlength{\tabcolsep}{4pt}
\begin{tabular}{lcccc}\toprule
Model & optical $2\times$ & digital $2\times$ & bilinear $2\times$ & JPEG \\ \midrule
DA-V2-S & 0.58 & 0.55 & 0.54 & 1.00 \\
Depth Pro & 0.83 & 0.75 & 0.82 & 1.00 \\
MoGe-2-S & 0.77 & 0.82 & 0.79 & 1.05 \\
Metric3D v2-S & 1.05 & 1.04 & 1.05 & 0.99 \\
\bottomrule\end{tabular}\end{table}

\paragraph{JPEG together with the zoom.} The JPEG control above compresses the unzoomed image. A camera that zooms
digitally saves the zoomed crop as a JPEG, and cropping a saved photo zooms into an image that is already
compressed. \Cref{tab:jpegzoom} runs both orders on the complete test sets. The audit itself never compresses a
zoomed image, so the zoom response of \cref{tab:audit} is not caused by compression; the question is whether
compression changes it. For four of the five models, compressing after the zoom moves depth by a factor of
\DavTwoSJpegAfterZoom{} to \MogeTwoSJpegAfterZoom{} relative to the plain zoom, about the size of the JPEG control
itself. Depth Pro is the exception: JPEG after the zoom lowers its depth by a further factor of
\DepthproJpegAfterZoom{} (\DepthproJzJpegOne{} for JPEG without the zoom), and the change sits in its focal
estimate, which moves by \DepthproJpegAfterZoomFocal. A digital zoom followed by compression, as on a phone,
therefore adds to Depth Pro's zoom error. Zooming into a JPEG adds more: \DepthproZoomOfJpeg{} for Depth Pro
(focal \DepthproZoomOfJpegFocal), \MogeTwoSZoomOfJpeg{} for MoGe-2-S and \UnidepthTwoSZoomOfJpeg{} for
UniDepthV2-S, for the last two largest on KITTI (\MogeTwoSZoomOfJpegKitti{} and \UnidepthTwoSZoomOfJpegKitti),
and less for DA-V2-S and Metric3D v2 (\DavTwoSZoomOfJpeg{} and \MetricThreedSZoomOfJpeg; DA-V2-S
\DavTwoSZoomOfJpegKitti{} on KITTI).

\begin{table}[t]\centering\small
\caption{JPEG (quality 75) together with a $2\times$ digital zoom, on the complete test sets (all four datasets pooled). Median depth ratio on the same pixels. Zoom: the plain zoom against the unzoomed image. JPEG $1\times$: compression alone. JPEG after zoom (a camera that zooms and saves a JPEG) and zoom of a JPEG (cropping a saved photo): against the plain zoom of the same image, so 1 means JPEG adds nothing.}\label{tab:jpegzoom}
\setlength{\tabcolsep}{3pt}
\begin{tabular}{lccccc}\toprule
Model & images & zoom & JPEG & JPEG after & zoom of \\
 & & $2\times$ & $1\times$ & zoom & a JPEG \\ \midrule
DA-V2-S & 2425 & 0.53 & 1.00 & 1.00 & 0.97 \\
Depth Pro & 2425 & 0.80 & 1.00 & 0.90 & 0.82 \\
MoGe-2-S & 2425 & 0.78 & 1.06 & 1.05 & 1.18 \\
Metric3D v2-S & 2425 & 1.04 & 1.00 & 1.00 & 0.98 \\
UniDepthV2-S & 2425 & 0.75 & 1.01 & 1.02 & 1.16 \\
\bottomrule\end{tabular}\end{table}

\paragraph{One change at a time.} \Cref{tab:factorial} collects these runs into a single factorial view of a
$2\times$ digital zoom: the changes that alter the camera (zoom, magnification of the content, shrinking the whole
image, an optical zoom with real detail) against those that keep it (removing the context around the crop,
losing detail by resizing). For the pixels-only DA-V2-S, magnification alone reproduces the zoom
(\DavTwoSCueMagnify{} against \DavTwoSZoomTwoFactor), shrinking moves depth the other way (\DavTwoSCueShrink), and
neither the lost context (\DavTwoSCueContext) nor the lost detail (\DavTwoSBlur\%) matters. Metric3D v2 separates
the camera from the pixels directly: the same pixels give \MetricThreedSZoomTwoFactor{} with the true $K'$ and
\MetricThreedSStaleKZoomTwoFactor{} with the stale $K$, which is the pixels-only response.

\begin{table*}[t]\centering\small
\caption{A $2\times$ digital zoom taken apart, one change at a time (numbers from the runs of \cref{tab:audit}, \cref{tab:cues} and \cref{tab:controls}). Entries: factor by which depth changes on the same pixels (1 = unchanged; apparent size alone: 0.5 for zoom and magnification, 2 for shrink). Zoom: $2^{-\alpha}$ from the audit. Crop only: the crop at its own place and size, original $K$ (context removed). Magnification: zoom against crop only. Shrink: the whole image $2\times$ smaller. Optical: the same field of view with real detail. Resize only: the blur control (detail lost, field of view and camera unchanged), absolute change. Only the factors that change the camera move depth; detail loss, compression and resampling do not.}\label{tab:factorial}
\setlength{\tabcolsep}{5pt}
\begin{tabular}{lcccccc}\toprule
& \multicolumn{4}{c}{camera changes} & \multicolumn{2}{c}{camera fixed} \\ \cmidrule(lr){2-5}\cmidrule(lr){6-7}
Model & zoom $2\times$ & magnification & shrink & optical $2\times$ & crop only & resize only \\ \midrule
DA-V2-S & 0.52 & 0.58 & 1.76 & 0.58 & 0.96 & 1.3\% \\
Depth Pro & 0.81 & 0.84 & 1.16 & 0.83 & 1.04 & 2.8\% \\
MoGe-2-S & 0.78 & 0.75 & 1.32 & 0.77 & 1.06 & 3.1\% \\
Metric3D v2-S, true $K'$ & 1.05 & -- & -- & 1.05 & -- & 1.6\% \\
Metric3D v2-S, stale $K$ & 0.53 & -- & -- & -- & -- & 1.6\% \\
\bottomrule\end{tabular}\end{table*}

\section{Real optical zoom}\label{sec:supp_srraw}

SR-RAW~\cite{zhang2019zoom} holds static scenes shot from a tripod with a Sony FE 24--240\,mm zoom lens at
seven focal lengths. The camera does not move, so the true depth of every point is the same in every shot and
no depth labels are needed. We use the camera JPEGs of the test set at a long side of 1024\,px and the EXIF focal
length; the shots with $s\le3.2$ against the widest one are compared, \srrawShots{} zoomed shots in
all \srrawSequences{} test sequences. As SR-RAW assumes, a zoomed shot shows the centred $1/s$ box of the wide shot; we
compare the median log depth over the inner 80\% of the zoomed shot with the median over the matching box of
the wide shot's prediction, which leaves a margin for the small shift of the lens centre. The same crop of the
wide shot, done digitally, is scored on the same scenes. Models that take the camera get $K$ from the EXIF
35\,mm-equivalent focal.

\begin{figure*}[t]
  \centering
  \maybefig{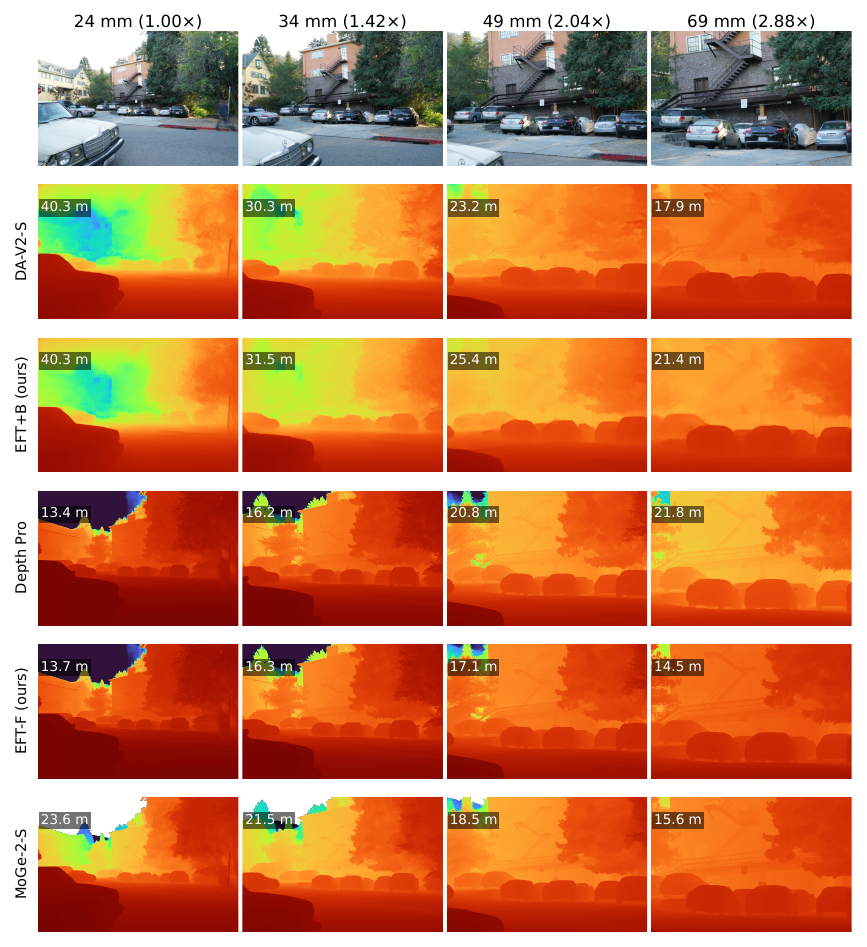}
  \caption{One SR-RAW sequence (the one with the median DA-V2-S $\alpha$, not hand-picked) shot at several
  focal lengths from a fixed tripod. The true depth does not change across a row; the number is the median
  predicted depth over the field of view of the longest shot, i.e.\ the same points in every column. Single
  sequences vary: here Depth Pro moves the scene away as the focal grows; \cref{tab:srraw} gives the medians.}
  \label{fig:supp_srraw}
\end{figure*}

\Cref{tab:srraw} gives the result. For DA-V2-S a real lens and a digital zoom are the same to the model
($\alpha=\DavTwoSSrrawOpticalAlpha$ and \DavTwoSSrrawDigitalAlpha, Wilcoxon $p=\DavTwoSSrrawP$ over
sequences): the error is not an artefact of resampling. Depth Pro keeps its depth under both
(\DepthproSrrawOpticalAlpha, \DepthproSrrawDigitalAlpha) and Metric3D v2 with the EXIF camera does too
(\MetricThreedSSrrawOpticalAlpha). MoGe-2-S and UniDepthV2-S shrink \emph{more} under the real lens than
under the digital zoom (\MogeTwoSSrrawOpticalAlpha{} against \MogeTwoSSrrawDigitalAlpha, and
\UnidepthTwoSSrrawOpticalAlpha{} against \UnidepthTwoSSrrawDigitalAlpha). EFT+B lowers $\alpha$ of DA-V2-S on
this unseen data to \DavTwoSEftbSrrawOpticalAlpha, less than on its training domains. \Cref{fig:supp_srraw}
shows the sequence whose DA-V2-S $\alpha$ is the median over all sequences.

\begin{table}[t]\centering\small
\caption{Real optical zoom (SR-RAW~\cite{zhang2019zoom}): the same static scene shot at several focal lengths ($s\le3.2$), against the same zoom done digitally from the widest shot. Median over sequences; ideal $\alpha=0$, $\beta=1$. The 392-px row is the frozen reference for EFT+B (same input size).}\label{tab:srraw}
\setlength{\tabcolsep}{2.5pt}
\begin{tabular}{lrcccc}\toprule
& & \multicolumn{2}{c}{$\alpha$} & \multicolumn{2}{c}{$\beta$} \\ \cmidrule(lr){3-4}\cmidrule(lr){5-6}
Model & seq. & optical & digital & optical & digital \\ \midrule
DA-V2-S & 50 & 0.76 & 0.76 & -- & -- \\
DA-V2-S, 392-px input & 50 & 0.74 & 0.71 & -- & -- \\
DA-V2-S + EFT+B (ours) & 50 & 0.59 & 0.59 & -- & -- \\
Depth Pro & 50 & \ensuremath{-0.08} & \ensuremath{-0.08} & 0.98 & 0.87 \\
Depth Pro + EFT-F (ours) & 50 & 0.04 & \ensuremath{-0.07} & 0.89 & 0.85 \\
MoGe-2-S & 50 & 0.38 & 0.26 & 0.58 & 0.69 \\
UniDepthV2-S & 50 & 0.62 & 0.39 & 0.22 & 0.53 \\
Metric3D v2-S & 50 & \ensuremath{-0.04} & \ensuremath{-0.03} & -- & -- \\
\bottomrule\end{tabular}\end{table}

\section{Fine-tuning curves}

\begin{figure}[t]
  \centering
  \maybefig{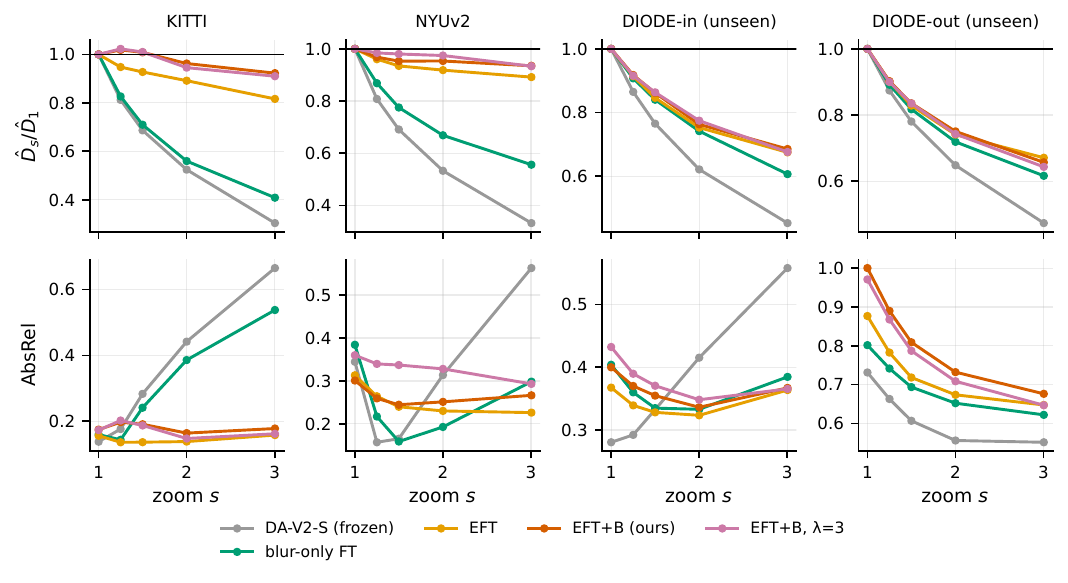}
  \caption{Depth ratio (top) and AbsRel (bottom) against the zoom factor for the frozen and fine-tuned
  models. AbsRel at different zooms is measured on different pixels (the crop); $\alpha$, not AbsRel, is
  the equivariance measure: on NYUv2 the frozen model over-estimates depth at $1\times$, so the shrink at
  moderate zoom happens to lower its error.}
  \label{fig:eft}
\end{figure}

\Cref{fig:eft} shows the depth ratio and the AbsRel of the frozen and fine-tuned DA-V2-S over the
whole zoom range, the curves behind the exponents of \cref{tab:eft}.

\section{Sanity check of the pipeline}

Projecting random 3D points with $K$, applying the crop-and-resize to the pixel coordinates,
and projecting with $K'$ agree to $<10^{-3}$\,px for all zooms and crop centres, and a constant
depth map survives the zoom/unzoom round trip exactly.

\section{Models and compute}

\paragraph{Checkpoints.} Hugging Face ids unless noted.
{\footnotesize\raggedright
\begin{itemize}\setlength{\itemsep}{0pt}
\item DA-V2: \texttt{depth-anything/Depth-Anything-V2-} \texttt{Metric-\{Indoor,Outdoor\}-\{Small,Base,Large\}-hf}
\item ZoeDepth: \texttt{Intel/zoedepth-nyu-kitti}
\item Depth Pro: \texttt{apple/DepthPro-hf}
\item MoGe-2: \texttt{Ruicheng/moge-2-\{vits,vitb,vitl\}-normal}
\item UniDepthV2: \texttt{lpiccinelli/unidepth-v2-} \texttt{\{vits14,vitb14,vitl14\}}
\item UniK3D: \texttt{lpiccinelli/unik3d-vitl}
\item DA3: \texttt{depth-anything/DA3NESTED-GIANT-} \texttt{LARGE-1.1}, \texttt{DA3METRIC-LARGE}, \texttt{DA3-LARGE-1.1}
\item VGGT: \texttt{facebook/VGGT-1B}
\item Metric3D v2: \texttt{metric3d\_vit\_\{small,large\}} (official torch hub entry)
\item Camera intervention (\cref{sec:camint}): MoGe-2-S and UniDepthV2-S with the checkpoints above, given the
true zoomed camera (MoGe-2: \texttt{fov\_x}; UniDepthV2: the \texttt{camera} argument of \texttt{infer})
\end{itemize}}

\paragraph{Compute.} The audits, EFT, EFT+B and EFT-F ran as Kaggle notebooks with one T4 GPU each. The cue
split and the EFT-S cache and fit ran as many parallel PyTorch jobs on the CPU cores
of a Kaggle TPU VM (the TPU itself was not used); the analysis runs on a laptop CPU. A check script confirms that
all models of one sample saw the same images, that every zoom row is complete, that the exact zoom factor matches
the nominal one, and that every SR-RAW shot has its optical and digital pair.

\end{document}